\documentclass[journal]{IEEEtai}

\usepackage[colorlinks,urlcolor=blue,linkcolor=blue,citecolor=blue]{hyperref}

\usepackage{color,array}

\usepackage{times}
\usepackage{soul}
\usepackage{url}
\usepackage{graphicx}
\usepackage{amsthm}
\usepackage{amsmath,amssymb}
\DeclareMathOperator{\E}{\mathbb{E}}
\usepackage{booktabs}
\usepackage{algorithm}
\usepackage{algorithmic}
\usepackage[switch]{lineno}
\usepackage{enumerate}
\usepackage{tabularx}
\usepackage{comment}
\usepackage{xcolor}
\usepackage{color, colortbl}
\usepackage{array, multirow, boldline}
\usepackage{adjustbox}
\usepackage{array}
\usepackage{breqn}
\usepackage{multirow}
\usepackage[raggedrightboxes]{ragged2e}
\usepackage{hyperref}
\usepackage{tabularx}
\usepackage{makecell} 
\hypersetup{
    colorlinks=true,
    linkcolor=black,
    filecolor=black,
    urlcolor=black
}

\begin{document}

\title{A Comprehensive Survey on Graph Summarization with Graph Neural Networks} 

\author{Nasrin Shabani, Jia Wu \IEEEmembership{Senior Member, IEEE}, Amin Beheshti, Quan Z. Sheng \\ Jin Foo, Venus Haghighi, Ambreen Hanif, Maryam Shahabikargar


\thanks{We acknowledge the Centre for Applied Artificial Intelligence at Macquarie University for funding this study.}

\thanks{Nasrin Shabani, Jia Wu, Amin Beheshti, Quan Z. Sheng, Venus Haghighi, Ambreen Hanif, and Maryam Shahabikargar 
are with the School of Computing, Macquarie University, NSW 2109, Australia. E-mails: \{nasrin.shabani@hdr, jia.wu, amin.beheshti, michael.sheng, eujin.foo@hdr, venus.haghighi@hdr, ambreen.hanif@hdr, maryam.shahabi-kargar@hdr\}mq.edu.au.}


\thanks{Corresponding authors: Nasrin Shabani and Amin Beheshti.}}


\markboth{IEEE Transactions on Artificial Intelligence, Vol. 00, No. 0, Month 2023}
{N. Shabani \MakeLowercase{\textit{et al.}}: A Comprehensive Survey on Graph Summarization With Graph Neural Networks}

\maketitle

\begin{abstract}
As large-scale graphs become more widespread, more and more computational challenges with extracting, processing, and interpreting large graph data are being exposed. It is therefore natural to search for ways to summarize these expansive graphs while preserving their key characteristics. In the past, most graph summarization techniques sought to capture the most important part of a graph statistically. However, today, the high dimensionality and complexity of modern graph data are making deep learning techniques more popular. Hence, this paper presents a comprehensive survey of progress in deep learning summarization techniques that rely on graph neural networks (GNNs). Our investigation includes a review of the current state-of-the-art approaches, including recurrent GNNs, convolutional GNNs, graph autoencoders, and graph attention networks. A new burgeoning line of research is also discussed where graph reinforcement learning is being used to evaluate and improve the quality of graph summaries. Additionally, the survey provides details of benchmark datasets, evaluation metrics, and open-source tools that are often employed in experimentation settings, along with a detailed comparison, discussion, and takeaways for the research community focused on graph summarization. Finally, the survey concludes with a number of open research challenges to motivate further study in this area. 
\end{abstract}

\begin{IEEEImpStatement}
Graph summarization is a key task in managing large graphs, which are ubiquitous in modern applications. In this article, we summarize the latest developments in graph summarization methods, offer a more profound understanding of these methods, and list source codes and available resources. The study covers a broad range of techniques, including both conventional and deep learning-based approaches, with a particular emphasis on GNNs. We aim to help the researchers develop a basic understanding of GNN-based methods for graph summarization, benefit from useful resources, and think about future directions.
\end{IEEEImpStatement}



\section{Introduction}
\IEEEPARstart{L}{arge} graphs are becoming increasingly ubiquitous. With the increasing amount of data being generated, large graphs are becoming more prevalent in modelling a variety of domains, such as social networks, proteins, the World Wide Web, user actions, and beyond. However, as these graphs grow in size, understanding and analyzing them is becoming more challenging. Additionally, performing fast computations with large graphs and visualizing the knowledge they can yield is also becoming more difficult. Many claim that faster and more effective algorithms are needed to overcome these obstacles~\cite{aggarwal2010managing,liu2018graph}. 
However, a growing cohort of researchers believe that summarization might hold the answer to this unyielding problem. Summarization not only helps existing algorithms to parse the data faster, it can also compress the data, reduce storage requirements, and assist with graph visualization and sense-making~\cite{vcebiric2019summarizing}.

\begin{figure}[t]
    \centering
    \includegraphics[width=0.5\textwidth]{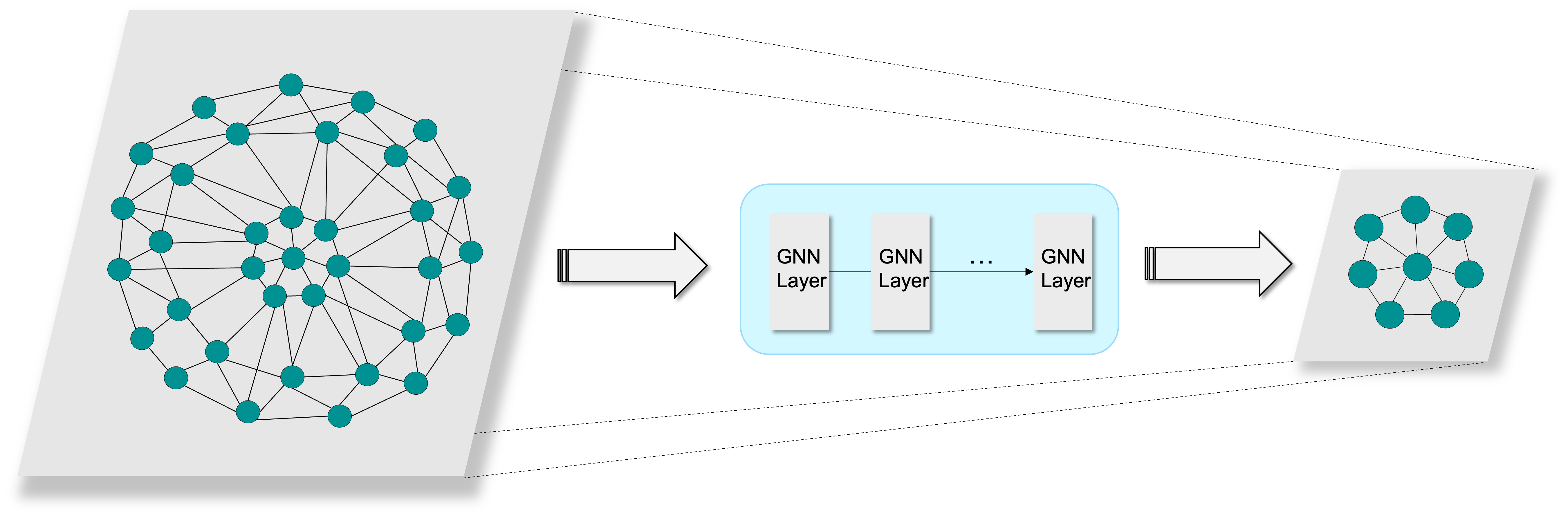}
    \caption{An example of a graph summarization process. Objects are removed from the original graph and replaced with fewer objects of the same type to result in a condensed representation of the original graph.}
    \label{fig:GS}
\end{figure}

Graph summarization is the process of finding a condensed representation of a graph while preserving its key properties~\cite{liu2018graph}. A toy example of a typical graph summarization process is shown in Figure~\ref{fig:GS}. The process includes removing the original graph’s objects and replacing them with fewer objects of the same type to produce a condensed representation of the original graph. 

Most traditional approaches to graph summarization involve using a conventional machine learning method or a graph-structured query, such as degree, adjacency, or eigenvector centrality, to find a condensed graphical representation of the graph~\cite{liu2018graph}. A popular summarization technique is to group structures in the input graph by aggregating the densest subgraphs~\cite{gibson2005discovering}. For example, the GraSS model~\cite{lefevre2010grass} focuses on accurate query handling and incorporates formal semantics for answering queries on graph structure summaries based on a random walk model, while Graph Cube~\cite{zhao2011graph} is a data warehousing model that integrates both network structure summarization and attribute aggregation. This model also supports OLAP queries on large multidimensional networks. 

Notably, clustering methods follow a similar approach to summarization, partitioning a graph into groups of nodes that can be further summarized. Most traditional graph clustering methods use conventional machine learning and statistical inference to measure the closeness of nodes based on their connectivity and structural similarities~\cite{kulis2005semi}. For instance, Karrer et al.~\cite{karrer2011stochastic} used a stochastic block model to detect clusters or communities in large sparse graphs. However, another method of graph summarization focuses more on node selection and identifying sparse graphs that can be used to derive smaller graphs~\cite{hu2013survey}. As an example, Doerr et al.~\cite{doerr2013metric} introduced a sampling method based on traversing a graph that begins with a collection of starting points, e.g., nodes or edges, and then adds to the sample pool depending on recent information about the graph objects. However, despite the popularity of these approaches in the past, they are very computationally-intensive. They also require a great deal of memory to store, making them unsuitable for today’s modern, complex, and large-scale datasets. 

Deep learning on graphs is a powerful computational technique for graphs of all sizes that continues to interest both academics and industry. Graph neural networks (GNNs) are the most successful paradigm among the deep learning techniques for graphs. Their multilayer deep neural networks are not only able to reduce the dimensionality of tasks, they also support high-speed calculations with large-scale graphs~\cite{zhou2020graph}. Several strategies that use a GNN as a summarization engine have been proposed over the last few years, and this line of research is expected to generate even more fruitful results in the future. These GNN-based methods have demonstrated promising performance with a diverse range of tasks, such as graph classification, node representation learning, and link prediction. Further, as research in this area continues to evolve, we can anticipate even more innovative and effective approaches to graph summarization that rely on GNNs.

\subsection{Existing Surveys on Graph Summarization }
Over the past decade, numerous reviews have been conducted that acknowledge the importance of graph summarization. They generally cover a range of topics, like graph partitioning, graph sampling, graph coarsening, and graph embedding, along with specific use cases for graph summarization, such as pattern discovery, community detection, fraud detection, and link prediction. Additionally, several comprehensive surveys on graph summarization techniques have been conducted based on scientific computing: 
\cite{liu2018graph,vcebiric2019summarizing,chen2022graph}. 
Yet only the most current work by Chen et al.~\cite{chen2022graph} compares the most recent machine learning techniques to traditional methods. 

There are also several surveys on graph sampling~\cite{zhang2021graph,liu2021sampling}, graph partitioning~\cite{ccatalyurek2022more,bhavsar2022graph}, and graph clustering~\cite{yue2022survey,lee2019review}. Other surveys focus on graph representation learning~\cite{xie2021survey,kazemi2020representation} and graph embedding~\cite{xu2021understanding,wang2022survey}. As these streams of research look to create low-dimension versions of a graph, they are strongly connected to the concept of graph summarization. However, although these studies provide in-depth analyses of how graph summarization techniques are being used in important and high-demand domains, GNN-based graph summarization methods are not their main area of focus. Consequently, these surveys do not provide a comprehensive and structured evaluation of all available techniques for graph summarization. 

\subsection{Contributions}
In this survey, we review developments in graph summarization with GNNs. Overall, we have aimed to provide researchers and practitioners with a comprehensive understanding of past, present, and future approaches to graph summarization using GNNs. Our specific contributions include:

\begin{itemize}
    \item \textbf{The first deep GNN-based graph summarization survey.} To our best knowledge, this paper is the first thorough survey that is devoted to graph summarization with GNNs.  Previous surveys have primarily concentrated on traditional graph summarization methods without considering deep learning techniques. As there is no existing specialized study on graph summarization with GNNs, this research aims to fill that gap and facilitate progress in the field through a detailed and structured survey.
    \item \textbf{Comprehensive review.} We present a comprehensive review of GNN-based graph summarization methods. We also highlight the benefits of different GNN techniques compared to traditional methods. Within each method, we present a review of the most notable approaches and their contributions to the field. 
    \item \textbf{Resources and reproducing existing results.} We have put together a set of resources that support graph summarization with GNNs, including cutting-edge models, standardized datasets for comparison, and reproducing existing publicly available implementations. Our goal is to provide a resource for those seeking to understand and apply GNNs for graph summarization.
    \item \textbf{Future directions.} We identify and discuss the open challenges and new directions that lie ahead. Through an exploration of the existing challenges and potential avenues for progress, we aim to guide future research and development in GNNs for graph summarization.
\end{itemize}

The majority of the papers we reviewed were published at prominent international conferences (e.g., SIGKDD, NeurIPS, ICLR, ICML, KDD, WWW, IJCAI, and VLDB) or in peer-reviewed journals (e.g, ACM, IEEE, Elsevier, and Springer) in the domains of artificial intelligence, big data, data mining, and  web services. In our investigations, we found that different fields referred to graph summarization using different terms, e.g., ``coarsening", ``reduction", ``simplification", ``abstraction", and ``compression". While these concepts were used relatively interchangeably, the terms coarsening, simplification, and summarization were generally more common. 

The remaining sections of this survey are structured as follows: Section~\ref{definitions} introduces the concept of graph summarization with primary definitions and background information. Section~\ref{development} provides an overview of the development of graph summarization. Section~\ref{GNN} outlines the recently developed GNN-based graph summarization methods. Section \ref{GRL} discusses graph reinforcement learning methods for graph summarization. Section \ref{resources} outlines the structure of widely adopted implementation resources. Finally, Section~\ref{future} explores a number of open research challenges that would motivate further study before the conclusion in Section~\ref{conclusion}.





\section{Definitions and background}
\label{definitions}

This section provides an overview of the key definitions and background information on graph summarization techniques.

\begin{itemize}
    \item \textit{Definition 1 (Graph)}: 
    Graph $G$ can be represented as a tuple $(V, E)$, where $V$ denotes the set of nodes or vertices $\{v_1, v_2, ..., v_n\}$, and $E$ represents the set of edges or links $E = \{e_{ij}\}^{n}_{i,j=1}$ connecting node pairs. The graph is represented by an $n \times n$ dimensional adjacency matrix $A = [a_{ij}]$, with $a_{ij}$ being 1 if the edge $e_{ij}$ is present in $E$ and 0 otherwise. If $a_{ij}$ is not equal to $a_{ji}$, the graph is directed; otherwise, it is undirected. When edges are associated with weights from the set $W$, the graph is called a weighted network; otherwise, it is an unweighted network. G is considered labeled if every edge $e \in E$ has an associated label. Additionally, if each node $v \in V$ has a unique label, the nodes are also labeled; otherwise, G is considered unlabelled.
    
    \item \textit{Definition 2 (Graph Summary)}: Given a graph $G$, a summary $G(V_S, E_s)$ is a condensed representation of $G$ that preserves its key properties. Graph summarization techniques involve either aggregation, selection, or transformation on a given graph and produce a graph summary as the output.
    
    \end{itemize}

As outlined in Definition 2, graph summarization approaches fall into three main categories: aggregation, selection, and transformation.
While selection approaches make graphs sparser by simply removing objects without replacing them, aggregation approaches replace those removed objects with similar objects only with fewer of them. For example, a supernode might replace a group of nodes. Similar to selection and aggregation, the transformation approaches also involve removing objects from the graph, but this time the objects removed are transformed into a different type of object, such as an embedding vector~\cite{jin2021graph}.

\textbf{Aggregation.} Aggregation is one of the most extensively employed techniques of graph summarization. Aggregation methods can be divided into two main groups: those that involve node grouping and those that involve edge grouping. Node grouping methods group nodes into supernodes, whereas edge grouping methods reduce the number of edges in a graph by aggregating them into virtual nodes. Clustering and community detection are examples of a grouping-based approach. Although summarizing graphs is not explicitly the primary objective of these processes, the outputs can be modified into non-application-specific summaries~\cite{liu2018graph}. 

\textbf{Selection.} There are two main groups of selection techniques:  sampling and simplification. While sampling methods focus on picking subsets of nodes and edges from the input graph~\cite{googleblog}, simplification or sparsification methods involve removing less important edges or nodes. In this way, they tend to resemble solutions to dimensionality reduction problems~\cite{hu2013survey}.

\textbf{Transformation.} Graph projection and graph embedding are two categories of this method. Generally, graph projection refers to the summarization techniques that transform bipartite graphs with various inter-layer nodes and edges into simple (single-layer) summarized graphs. Conversely, graph embedding refers to the techniques that transform a graph into a lower dimensional representation while preserving the original graph’s topology~\cite{interdonato2020multilayer}.

\section{Graph Summarization: An Evolution}
\label{development}

\begin{figure*}[t]
    \centering
    \includegraphics[width=1\textwidth]{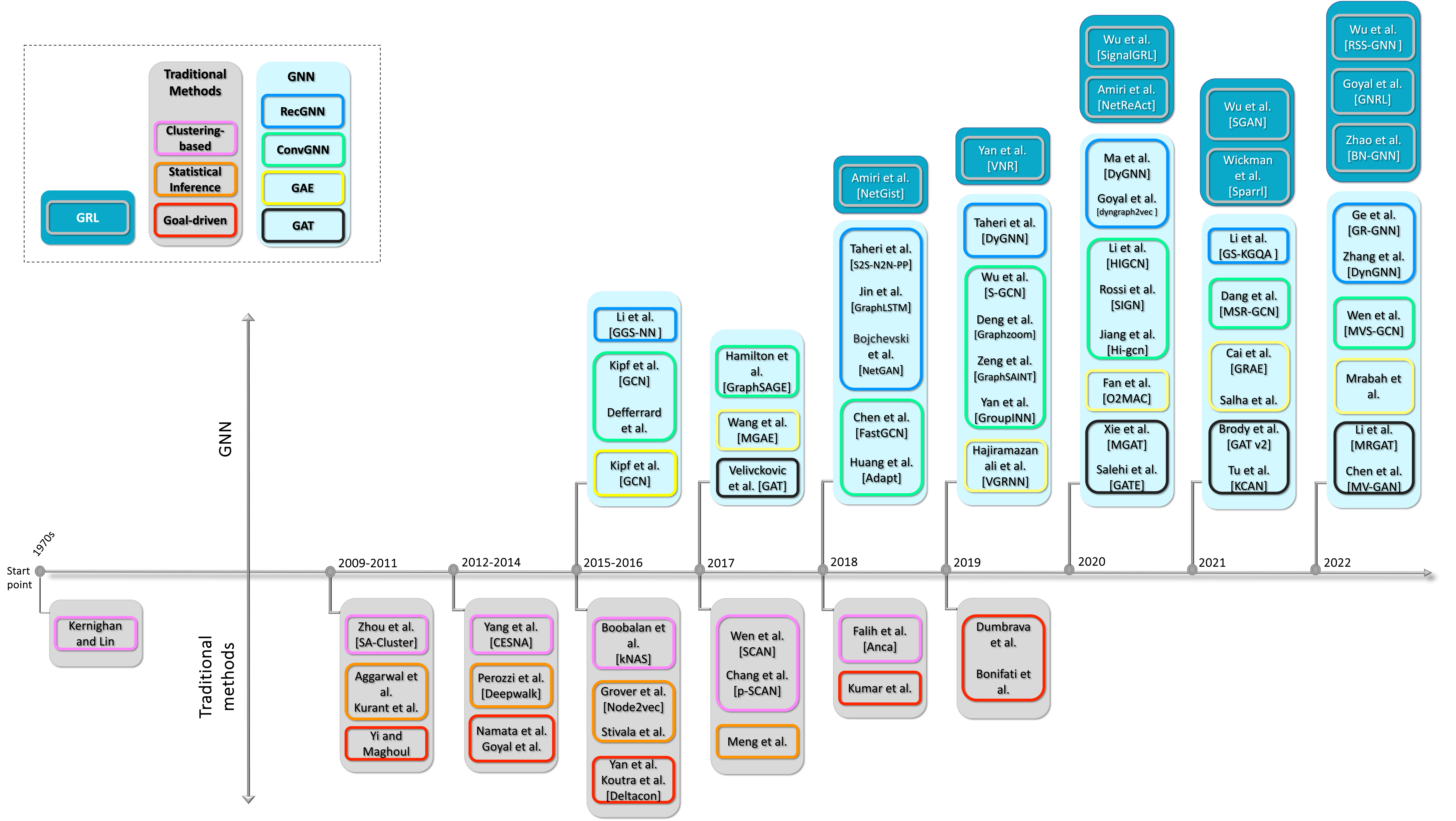}
    \caption{A timeline of graph summarization and reviewed techniques.}
    \label{fig:timleline}
\end{figure*}

Graph summarization has been playing an important role in areas such as network analysis, data mining, machine learning, and visualization for some time. The evolution of graph summarization is illustrated in Figure \ref{fig:timleline}, which shows how it has progressed from traditional computing methods to multi-layer GNNs. 
This section briefly overviews the three different traditional methods within this field and explains the advantages of GNN techniques over traditional ones.

\subsection{Clustering-based Approaches}
Graph clustering can be thought of as a graph summarization technique since it involves grouping together nodes in a graph that are similar or related and, in so doing, the complexity and size of the original graph are reduced. In simpler terms, graph clustering provides a way to compress or summarize a large and complex graph into a smaller set of clusters, each of which captures some aspect of the structure or function of the original graph~\cite{aggarwal2010managing}. Graph summarization techniques using clustering can be classified into three main categories: structural-based, attribute-based, and structural-attribute-based approaches. The latter, combining both structural and attribute information, is considered the most effective~\cite{chodrow2021generative}. 
For example, Boobalan et al.~\cite{boobalan2016graph} proposed a method called k-Neighborhood Attribute Structural Similarity (k-NASS) that incorporates both structural and attribute similarities of graph nodes. This method improves clustering accuracy for complex graphs with rich attributes. However, clustering large graphs with many attributes remains challenging due to high memory and computational requirements.

\subsection{Statistical Inference}
Statistical inference techniques for graph summarization simplify the complexity of the original graph while preserving its significant characteristics. These techniques fall into two groups: pattern mining and sampling. Pattern mining identifies representative patterns or subgraphs in the graph to create a condensed summary. On the other hand, sampling randomly selects a subset of nodes or edges from the graph and estimates the properties of the entire graph based on this subset. One example of a sampling technique is Node2vec~\cite{grover2016node2vec}, which generates random sequences of nodes within a graph, known as walks, to create a graph summary. Various sampling techniques, such as random sampling~\cite{grover2016node2vec}, stratified sampling~\cite{kurant2011walking}, and snowball sampling~\cite{stivala2016snowball}, can be used for graph summarization. Each technique has its advantages and disadvantages, and the choice depends on the specific problem and data being addressed.

\subsection{Goal-driven}
Goal-driven techniques for graph summarization involve constructing a graph summary that is tailored to a specific application or task. They are a powerful tool for capturing specific features or relationships in a graph that are relevant to a specific application or task. By optimizing the graph summary to a specific goal, it is possible to create a more effective and efficient summary that can be used to derive better insights and make better decisions~\cite{hajiabadi2022efficient}. Significant goal-driven techniques for graph summarization include utility-driven and query-driven techniques. Utility-driven techniques aim to summarize large graphs while preserving their essential properties and structure to maximize their usefulness for downstream tasks. Human reviewers evaluate the utility of the summary against specific tasks like node classification and link prediction~\cite{kumar2018utility}. Query-driven techniques summarize graphs by identifying relevant subgraphs or patterns using queries in a query language. The resulting subgraph that matches the query becomes a building block for the graph summary, supporting the target downstream task~\cite{dumbrava2019approximate}. The choice of the goal-driven summarization technique depends on the specific goals of the analysis, as some techniques may preserve global properties, while others may capture local structures. It also depends on available computational resources and the complexity and size of the original graph.

\begin{figure*}[h]
    \centering
    \includegraphics[width=1\textwidth]{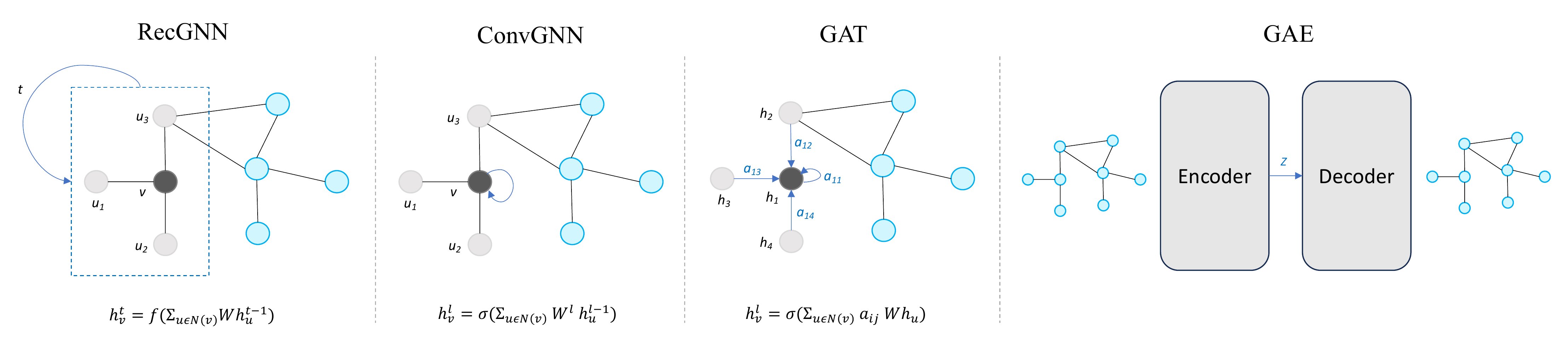}
    \caption{GNN Architechtures \cite{wu2020comprehensive,dong2023graph}.}
    \label{fig:GAN_Arch}
\end{figure*}

\subsection{Why GNNs for Graph Summarization?}
In recent times, deep learning has gained significant prominence and is now considered one of the most effective forms of AI due to its high accuracy. Conventional deep learning methods have shown that they perform extremely well with Euclidean data (e.g., images, signals, and text), and now there are a growing number of applications in non-Euclidean domains (e.g., graphs and manifold structures). As a deep learning approach, GNNs are multi-layer neural networks that learn on graph structures to ultimately perform graph-related tasks like classification, clustering, pattern mining, and summarization~\cite{wu2020comprehensive}. 

As mentioned, traditional graph summarization approaches are mostly based on conventional machine learning or graph-structured queries, such as degree, adjacency, and eigenvector centrality, where the aim is to find a condensed graphical representation of the whole graph~\cite{zhao2011graph}. However, the pairwise similarity calculations involved in these approaches demand a considerably high level of computational power. The explicit learning capabilities of GNNs skirt this problem. Additionally, powerful models can be built from even low-dimensional representations of attributed graphs~\cite{cai2018comprehensive}. Unlike standard machine learning algorithms, with a GNN, there is no need to traverse all possible orders of the nodes to represent a graph. Instead, GNNs consider each node separately without taking the order of the input nodes into account. This avoids redundant computations. 

The major advantage of GNN models for graph summarization over traditional methods is the ability to use low-dimensional vectors to represent the features of large graphs~\cite{liu2020deep}. Additionally, the message-passing mechanism used by GNNs to communicate information from one node to the next has been the most successful learning framework for learning the patterns and neighbours of nodes and the sub-graphs in large graphs~\cite{zhou2020graph}. It is also easy to train a GNN in semi- or unsupervised way to  aggregate, select, or transform graphs into low dimensional representations~\cite{wu2022graph}. In this regard, recent successes in graph summarization with GNNs point to promising directions for new research. For example, the GNN models developed by Brody et al.~\cite{brody2021attentive} and Goyal et al.~\cite{goyal2020dyngraph2vec} represent dynamic graphs in low dimensions, providing a good foundation for popularizing GNNs into more complex dynamic graphs. 

\section{Graph Summarization with GNNs}
\label{GNN}

This section provides an overview of recent research into graph summarization with GNNs. Each subsection covers four main categories of approach – these being: Recurrent Graph Neural Networks (RecGNNs), Convolutional Graph Neural Networks (ConvGNNs), Graph Autoencoders (GAEs), and Graph Attention Networks (GATs). Four different types of GNN models are shown in Figure \ref{fig:GAN_Arch}. Within each subsection, we first provide a brief introduction about the architecture of the GNN model and then review the most notable approaches and the contributions each has made to the field. At the end of each subsection, we provide a comprehensive summary of the key features of representative GNN-based approaches. We present a comparative analysis in Tables~\ref{tab:recgnn_comparison},~\ref{tab:convgnn_comparison},~\ref{tab:gae_comparison}, and~\ref{tab:gat_comparison} for RecGNN, ConvGNN, GAE, and GAT architectures, respectively. The tables include comparisons of evaluation methods, performance metrics, training data, advantages, and limitations across the different models.

\subsection{RecGNN-based Approaches}
RecGNNs are early implementations of GNNs, designed to acquire node representations through a generalized recurrent neural network (RNN) architecture. Within these frameworks, information is transmitted between nodes and their neighbours to reach a stable equilibrium~\cite{wu2020comprehensive,huang2019inductive}. A node's hidden state is continuously updated by 
\begin{equation}
\label{equ:RecGNNs}
h_{v}^{t} = f(\sum_{u \in N(v)} W h_{u}^{t-1})
\end{equation}
where \(h_{v}^{t}\) is the hidden state of node \(v\) at time \(t\), which represents the information learned by the RecGNN about node \(v\) at a specific time step in the dynamic graph sequence. \(N(v)\) denotes the set of neighboring nodes of node \(v\) in the graph, providing the context and connectivity information for node \(v\) within the graph structure. \(h_{u}^{t-1}\) is the hidden state of neighboring node \(u\) at time \(t-1\), which contributes to the update of node \(v\)'s hidden state at the previous time step, reflecting the influence of neighboring nodes on \(v\)'s representation. \(W\) is the weight matrix used for aggregating the hidden states of neighboring nodes.

Here, \(f(\cdot)\) is usually a simple element-wise activation like ReLU, tanh, or sigmoid. This simple activation is typically used to introduce non-linearity and capture complex patterns in the node representations.
However, this can be replaced by recurrent update functions, which use gated mechanisms like LSTM (Long Short-Term Memory)~\cite{Hochreiter1997lstm} or GRU (Gated Recurrent Units)~\cite{cho2014learning} cells. In this case, each node's hidden state update would be computed using an LSTM or GRU cell, which is more complex and sophisticated than simple element-wise activation functions. These cells determine how the hidden state of node \(v\) at time \(t\) is updated based on its current and previous hidden states and the information from its neighboring nodes, enabling the model to capture temporal dependencies in the dynamic graph-structured data~\cite{lai2021mgrnn}. 
 
RecGNN-based approaches for graph summarization mostly focus on graph sampling and embedding by generating sequences from graphs and embedding those sequences into a continuous vector space at lower dimensions. In the following, we will first briefly introduce LSTM and GRU architectures and then delve into the graph summarization approaches that are built upon their respective structures. 

\begin{table*}[t]
\centering
\caption{Comparative analysis of selected RecGNN-based approaches for Graph Summarization.}
\label{tab:recgnn_comparison}
\begin{adjustbox}{max width=\textwidth}
\begin{tabular}{m{1.5cm}m{1.5cm}m{1cm}m{2.5cm}m{2cm}m{1.5cm}m{4cm}m{4cm}}
\toprule
\textbf{Ref.} & \textbf{Model Name} & \textbf{Approach} & \textbf{Evaluation} & \textbf{Performance Metrics} & \textbf{Training Data} & \textbf{Advantages} & \textbf{Limitations} \\
\midrule
Taheri et al.~\cite{taheri2018learning} & S2S-N2N-PP & LSTM & Node Classification & Accuracy & Labeled, Unlabeled & Robust performance, capturing global structure, unsupervised learning. & Limited performance with Weisfeiler-Lehman labels, sensitive to noise. \\
Jin et al.~\cite{jin2018learning} & GraphLSTM &LSTM & Graph Classification & Accuracy & Labeled  & Incorporation of long-term dependencies, effective node representations. & Dependency on pretrained embeddings, sensitivity to node ordering, lack of discussion on scalability. \\
Bojchevski et al.~\cite{bojchevski2018netgan} & NetGAN &LSTM & Link Prediction & Link Prediction Accuracy & Labeled & Generative approach, scalability, unsupervised learning. & practical limitations with massive graphs, training instability due to use of GANs. \\
Wu et al.~\cite{wu2020graph} & GSGAN &LSTM & Community Detection & ARI & Labeled & Flexibility in sparsification, handling large data by using random walks, accurate noise filtering.  & Limited guarantee for analysis, adding artificial edges (contradicts with the graph summariziation's aim), lack of comparison with other models. \\
Ma et al.~\cite{ma2020streaming} & DyGNN &LSTM & Node Classification, Link Prediction & MRR, Recall & Labeled, Unlabeled & Incorporation of temporal information, consideration of varied influence. & Limited label information, no parameter analysis for hyper parameters. \\
Khoshraftar et al.~\cite{khoshraftar2019dynamic} & LSTM-Node2vec & LSTM & Node Classification, Link Prediction, Anomaly Detection & AUC, F1-score & Labeled & Incorporates temporal information, outperforms state-of-the-art methods, preserves long-term dependencies. & Fixed length of history, Memory intensive, no attention mechanism. \\
Li et al.~\cite{li2015gated} & GGS-NNS &GRU & Node Classification & Accuracy & Labeled  & Capturing temporal information, Reasonable scalability, end-to-end learning & Limited global information, complexity for large graphs, task-specific architecture. \\
Taheri et al.~\cite{taheri2019learning} & DyGGNN &GRU & Graph Classification & Accuracy & Labeled & Capturing topological and temporal features, Sequence-to-sequence architecture, unsupervised. & Not scalable to very large graphs, Limited evaluation, complexity and computational cost. \\
Ge et al.~\cite{ge2022gr} & GR-GNN & GRU & Graph Classification, Node Classification & Micro F1-score & Labeled & Deep feature extraction, robustness and universality, computational efficiency. & Limited scope of evaluation, limited extraction of deep chain-dependent features. \\

\bottomrule
\end{tabular}
\end{adjustbox}
\end{table*}

\subsubsection{LSTM-based Approaches}
LSTMs are a class of RNNs that use a sequence of gates to concurrently model long- and short-term dependencies in sequential data~\cite{medsker2001recurrent}. The modified architecture of an LSTM to handle large graph-structured data is known as a GraphLSTM~\cite{liang2016semantic}. Typically, the input to the model consists of a sequence of either graph nodes or edges, which are processed in order using LSTM units. At each time step, the model updates its internal state based on the input node or edge and the previous state, as shown in Equation \ref{equ:LSTM}. 
\begin{equation}
\label{equ:LSTM}
h_v^t = LSTM(\sum_{u \in N(v)} W h_{u}^{t-1})
\end{equation}
The cell is composed of multiple gates, and its operation can be described as follows~\cite{liang2016semantic}: 

\begin{equation}
f_v^t = \sigma(W_f \times [h_v^{t-1}, x_v^t] + b_f)
\end{equation}
\begin{equation}
i_v^t = \sigma(W_i [h_v^{t-1}, x_v^t] + b_i)
\end{equation}
\begin{equation}
o_v^t = \sigma(W_o [h_v^{t-1}, x_v^t] + b_o)
\end{equation}
\begin{equation}
C_v^t = tanh(W_C [h_v^{t-1}, x_v^t] + b_C)
\end{equation}
\begin{equation}
c_v^t = f_v^t \times c_v^{t-1} + i_v^t \times C_v^t
\end{equation}

In the given equations, the variables \(f_v^t\), \(i_v^t\), and \(o_v^t\) serve as the forget, input, and output gates, respectively. The forget gate plays a crucial role in determining what information to retain or discard from the cell state (long-term memory) for node \(v\) at time \(t\). On the other hand, the input gate is responsible for determining which new information should be stored in the cell state for the node. Also, the output gate regulates what information is to be outputted from the current cell state. Finally, the hidden state for node $v$ at time $t$ is updated using the output gate and the cell state as follows:

\begin{equation}
\label{equ:LSTM-ht}
    h_v^t = o_v^t \times tanh(c_v^t)
\end{equation}

The cell state \(c_v^t\) represents the memory cell of the LSTM network at time \(t\). It acts as a long-term memory, capable of storing information over extended sequences. To calculate \(c_v^t\), the new candidate value \(C_v^t\) is first calculated and then the cell state for node $v$ at time $t$ is updated using the forget gate, input gate, and the new candidate values. The gates take into consideration the previous hidden state \(h_v^{t-1}\) of node \(v\) at time \(t-1\) and the current input feature vector \(x_v^t\). Moreover, \(W_x\) and \(b_x\) are the weights and biases, respectively, associated with each of the gates.

By repeating this process over multiple time steps, the model captures the dependencies in the graph and generates a final hidden state that summarizes all the information contained in the entire graph. This makes it possible to preserve a record of both the previous inputs and the structure of the graph. After processing all the nodes and edges, the final state of the graph LSTM is used as the summary representation of the graph. This summary vector can then be used as input to downstream machine learning models or as a feature for other graph analysis tasks~\cite{zhang2022dynamic}. 

LSTM-based approaches for graph summarization have proven to be effective with a range of tasks, such as graph clustering, graph classification, and link prediction. For example, Taheri et al.~\cite{taheri2018learning} leveraged various graph-LSTM-based methods to generate sequences from a graph structure, including breadth-first search, random walk, and shortest path. Finally, they trained a graph LSTM to embed those graph sequences into a continuous vector space. The hidden states at time step $t$ of the encoder are as follows:
\begin{equation}
\label{equ:taheri2018_1}
h_t^{enc} = LSTM_{enc} (x_t, h_{t-1}^{enc})
\end{equation}
where $x_t$ is the input vector at time $t$ and $h_t^{enc}$ denotes the hidden state at time step t in a given trained encoder $LSTM_{enc}$. Similarly, the hidden state at time step $t$ of the decoder is defined in Equation \ref{equ:taheri2018_2}.
\begin{equation}
\label{equ:taheri2018_2}
h_t^{dec} = LSTM_{dec} (x_{t-1}, h_{t-1}^{dec})
\end{equation}

Bojchevski et al. \cite{bojchevski2018netgan} proposed NetGAN, a recurrent-based model to capture the underlying structural information that leads to the learning of meaningful network embeddings. NetGAN leverages the power of generative models to learn a compact representation of a graph via random walks, enabling more efficient processing of large graphs while preserving their essential features. As a variant, Wu et al.~\cite{wu2020graph} modified the NetGAN model to form a new social network with artificial edges that is suitable for community detection.

Jin et al.~\cite{jin2018learning} also developed an approach to learning representations of graphs based on a graph LSTM. Here, graph representations of diverse sizes are encoded into low-dimensional vectors. Li et al.~\cite{li2021improving} proposed a graph summarization technique that uses a graph LSTM and a ConvGNN to improve question answering with knowledge graphs. In this approach, the questions, entities, and relations are represented as vectors with very few dimensions, but the key properties of the relations are well preserved. 

Several studies have also focused on evolving node patterns in dynamic graphs. For instance, Zhang et al.~\cite{zhang2022dynamic} introduced an LSTM-based approach, a one-stage model called DynGNN. The model embeds an RNN into a GNN model to produce a representation in compact form. Khoshraftar et al.~\cite{khoshraftar2019dynamic} presented a dynamic graph embedding method via LSTM to convert a large graph into a low-dimensional representation. The model captures temporal changes with LSTM using temporal walks and then transfers the learned parameters into
node2vec~\cite{grover2016node2vec} to incorporate the local structure of each graph. Similarly, Ma et al.~\cite{ma2020streaming} introduced a dynamic RecGNN model that relies on a graph LSTM to model the dynamic information in an evolving graph while reducing the graph’s dimensionality and learning manifold structures. Node information is continuously updated by: recording the time intervals between edges; recording the sequence of edges; and coherently transmitting information between nodes. Another work by Goyal et al.~\cite{goyal2020dyngraph2vec} also presents a method for learning temporal transitions in dynamic graphs. This framework is based on a deep architecture that mainly consists of dense and recurrent layers. Model size and the number of weights to be trained can be a problem during training, but the authors overcome this issue with a uniform sampling of nodes.  

\subsubsection{GRU-based Approaches}
GRUs are a variant of graph LSTMs that include a gated RNN structure and have fewer training parameters than a standard graph LSTM. The key distinction between a GRU and an LSTM is the number of gates in each model. GRU units are less complex with only two gates, ``reset" and ``update"~\cite{zarzycki2022advanced}.

\begin{equation}
    \label{chebshev}
    r_v^t = \sigma(W_r [h_v^{t-1}, x_v^t] + b_r)
\end{equation}

\begin{equation}
    \label{chebshev}
    z_v^t = \sigma(W_z [h_v^{t-1}, x_v^t] + b_z)
\end{equation}

\begin{equation}
    \label{chebshev}
    C_v^t = tanh(W_C [r_v^t \times h_v^{t-1}, x_v^t] + b_C)
\end{equation}

\begin{equation}
    \label{chebshev}
    h_v^t = (1 - z_v^t) \times h_v^{t-1} + z_v^t \times C_v^t
\end{equation}

In these equations, $r_v^t$ is a reset gate and $z_v^t$ is update gate. $h_v^{t-1}$ is the output of the model at the previous time step. Similar to LSTM, $C_v^t$ computes the new candidate value, and $h_v^t$ updated hidden state for node $v$ at time $t$ using the update gate and the new candidate values. $b_x$, $W_x$ are biases and weights for respective gates.

Again, by repeating this process over several time steps, the model learns the dependencies that exist between the nodes in the graph, allowing it to construct a final hidden state that summarizes all the graph’s information. The adaptability of GRU cells to capture temporal dependencies within graph-structured data allows for effective information aggregation and context modeling, making GRU-based methods a promising choice for summarizing complex graph-structured information. For instance, Taheri et al.~\cite{taheri2019learning} proposed the DyGrAE model, which is able to learn the structure and temporal dynamics of a dynamic graph while condensing its dimensions. A GRU model captures the graph's topology, while an LSTM model learns the graph’s dynamics. Ge et al.~\cite{ge2022gr} developed a gated recursive algorithm that not only solves some node aggregation problems but also extracts deeply dependent features between nodes. The resulting model, called GR-GNN, is based on a GRU which performs the aggregation and structure. Li et al.’s GRU model~\cite{li2015gated} encodes an input graph into a fixed-size vector representation, which is then fed into a sequence decoder to generate the summary as the output. The model effectively captures the structural information and dependencies among the nodes and edges in the input graph, which is crucial for producing a coherent and informative graph summary.


\begin{table*}[t]
\centering
\caption{Comparative analysis of selected ConvGNN-based approaches for Graph Summarization.}
\label{tab:convgnn_comparison}
\begin{adjustbox}{max width=\textwidth}
\begin{tabular}{m{1.5cm}m{1.5cm}m{1cm}m{2.5cm}m{2cm}m{1.5cm}m{4cm}m{4cm}}
\toprule
\textbf{Ref.} & \textbf{Model Name} & \textbf{Approach} & \textbf{Evaluation} & \textbf{Performance Metrics} & \textbf{Training Data} & \textbf{Advantages} & \textbf{Limitations} \\
\midrule
Kipf et al.~\cite{kipf2016semi} & GCN &Spectral & Node Classification & Accuracy & Labeled & Overcoming limitations of graph laplacian regularization, less complex and better scalability, improved predictive performance. & Memory requirement, limited to directed edges, does not support edge features, limiting assumptions.\\
Wu et al.~\cite{wu2019simplifying} & SGC &Spectral & Node Classification & Accuracy, Micro F1-score & Labeled & Ease of implementation, applicability to large-scale graphs, memory efficiency. & Limited expressiveness, loss of hierarchical representations, limitations in handling complex graphs due to its simplified nature.\\
Deng et al.~\cite{deng2019graphzoom} & GraphZoom &Spectral & Node Classification & Accuracy, Micro F1-score & Labeled & Applicable to different types of graphs, preservation of local and global structure, scalable to huge graphs. & Lack of node label preservation, limited support for large graphs. \\
Rossi et al.~\cite{rossi2020sign} & SIGN&Spectral & Node Classification & Accuracy & Labeled & Robustness to graph irregularities, general applicability (can be applied to various graph-based tasks), faster training and inference. & Limited to undirected graphs, dependent on the choice of operator combinations.\\
Jiang et al.~\cite{jiang2020hi} & Hi-GCN & Spectral & Graph Classification & Accuracy, AUC, Sensitivity and Specificity & Labeled & Hierarchical graph convolution, contribution to neuroscience, consideration of correlation. & Complex optimization, limited comparison with prior works.\\
Hamilton et al.~\cite{hamilton2017inductive} & GraphSAGE &Spatial & Node Classification & Micro F1-score & Labeled & Scalability, 
inductive learning, flexibility in using aggregation strategies. & Assumption of homophily, limited global context, over-smoothing, hyperparameter sensitivity.\\
Chen et al.~\cite{chen2018fastgcn} & FastGCN &Spatial & Node Classification & Micro F1-score & Labeled, Unlabeled & Ease of sampling implementation, retains model accuracy despite using importance sampling to speed up training. & Limited comparison with the state-of-the-art, opportunities to reduce sampling variance remain.\\
Zeng et al.~\cite{zeng2019graphsaint} & GraphSAINT &Spatial & Node Classification & Accuracy, Micro F1-score & Labeled & Efficient training on large graphs by introducing "neighbor explosion", low training complexity. & Sampling overhead, pre-processing requirements, limited generalization to unseen graphs. \\
Yan et al.~\cite{yan2019groupinn} & GroupINN &Spatial & Node Classification & Accuracy, Micro F1-score & Labeled & Provides interpretability, parameter reduction, capturing complex relationships. & Limited application, dependency on data quality, lack of generalizability to different datasets or tasks outside the scope of the neuroscience domain. \\
Wen et al.~\cite{wen2022mvs} & MVS-GCN &Spatial & Graph Classification & Accuracy, AUC, Sensitivity and Specificity & Labeled & Multi-view brain network embedding, interpretability, effective graph structure learning. & Limited contribution of negative connectives, influence of hyperparameters, challenges in interpreting complex brain networks. \\
\bottomrule
\end{tabular}
\end{adjustbox}
\end{table*}

\subsection{ConvGNN-based Approaches}  
The general idea of ConvGNN-based approaches is to generalize the CNNs on graph-structured data~\cite{wu2020comprehensive}. The primary distinction between a ConvGNN and a RecGNN is the way information is propagated. While ConvGNNs apply various weights at each timestep, RecGNNs apply the same weight matrices in an iterative manner until an equilibrium is reached~\cite{zhang2019graph}. 

In other words, ConvGNN models are a form of neural network architecture that supports graph structures and aggregates node information from the neighbourhood of each node in a convolutional manner. ConvGNN models have demonstrated a strong expressive capacity for learning graph representations, resulting in superior performance with graph summarization~\cite{zhang2019graph}. 

ConvGNN-based approaches fall into two categories: spectral-based and spatial-based methods~\cite{wu2020comprehensive}. 

\subsubsection{Spectral-based Approaches}

Spectral-based methods describe graph convolutions based on spectral graph theory and graph signal filtering. In spectral graph theory, the multiplication of the graph with a filter (the convolution) is defined in a Fourier domain~\cite{levie2018cayleynets}. 

Although the computation contains well-defined translational properties, it is relatively expensive, and the filters are not generally localized. Since the level of complexity grows with the scale of the graphs, one solution is to only check a limited number of neighbours using Chebyshev’s theory~\cite{hammond2011wavelets}. The Chebyshev polynomials \(T_k(x)\) are defined recursively as:

\begin{equation}
    \label{chebshev}
    T_k(x) = 2x T_{k-1}(x) - T_{k-2}(x)
\end{equation}
where $T_0(x) = 1$ and $T_1(x) = x$. Here, $x$ represents the variable of the Chebyshev polynomial. $k$ is a non-negative integer that determines the degree of the polynomial which is the order of the Chebyshev polynomial. 
It is a polynomial function of degree \(k\), and its value depends on the values of \(x\) and \(k\) as defined by the recurrence relation mentioned in Equation~\ref{chebshev}.

This theory has led to several studies that explore the idea of applying approximation to the spectral graph convolution. For example, Defferrard et al.~\cite{defferrard2016convolutional} generalized the CNNs to graphs by designing localised convolutional filters on graphs. Their main idea was to leverage the spectral domain of graphs and use Chebyshev polynomial approximation to efficiently compute localized filters as follows: 

\begin{equation}
    \label{defferrard}
    Z = \sum_{k=0}^{K-1} \theta_k T_k(\widetilde{L}) X
\end{equation}
where $L$ represents a graph Laplacian, $X$ is the node feature matrix, $\theta$ is the graph convolutional operator with a filter parameter, $\widetilde{L}$ is a scaled Laplacian defined as $\widetilde{L} = 2L/I - I$, and $K$ is the order of the Chebyshev polynomial approximation. The $\theta_k$ parameters are the learnable filter coefficients. They also introduced a graph summarization procedure that groups similar vertices together and a graph pooling procedure that focuses on producing a higher filter resolution. This work has been used as the basis of several studies that draw on Chebyshev polynomials to speed up convolution computations. 

As a variant, Kipf et al.~\cite{kipf2016semi} introduced several simplifications to the original framework to improve the model’s classification performance and scalability to large networks. These simplifications formed the basis of the GCN (Graph Convolutional Network) model, which has since become a popular choice for various graph-related tasks. Given an undirected graph with an adjacency matrix $A$, they computed the normalized graph Laplacian $\widetilde{L}$ as follows:

\begin{equation}
    \label{kipf-gcn}
    \widetilde{L} = I - D^{-1/2}AD^{-1/2}
\end{equation}
where \(I\) is the identity matrix, and \(D\) is the diagonal degree matrix, with \(D_{ii}\) representing the sum of the weights of the edges connected to node \(i\).

In this work, instead of directly computing graph convolutions using high-order Chebyshev polynomials, as done in the previous work by Defferrard et al.~\cite{defferrard2016convolutional}, Kipf et al. proposed using a simple first-order approximation of graph filters. They defined the graph convolution operation as~\cite{kipf2016semi}:

\begin{equation}
    \label{kipf-gcn1}
    H^{(l+1)} = \sigma(\widetilde{D}^{-1/2}\widetilde{A}\widetilde{D}^{-1/2}H^{(l)}W^{(l)})
\end{equation}
where $H^{(l)}$ represents the hidden node features at layer $l$. $W^{(l)}$ is the weight matrix for the layer $l$. $\widetilde{A} = A + I$ is the adjacency matrix with self-loops added. And, $\widetilde{D}$ is the diagonal degree matrix of $\widetilde{A}$. Here, the normalized graph Laplacian \( \widetilde{D}^{-1/2} \widetilde{A} \widetilde{D}^{-1/2} \) is used to aggregate information from neighboring nodes. Hence, the propagation can be written as follows:

\begin{equation}
\label{kipf-gcn2}
  h_i^{(l+1)} = \sigma \left( \sum_{j \in \mathcal{N}_i} \frac{1}{\sqrt{\widetilde{D}_{ii} \widetilde{D}_{jj}}} \cdot h_j^{(l)} \cdot W^{(l)} \right) 
\end{equation}
where \( h_i^{(l)} \) is the feature vector of node \( i \) at layer \( l \). \( \mathcal{N}_i \) is the set of neighbors of node \( i \) in the graph. \( W^{(l)} \) is the weight matrix for the \( l \)-th layer. \( \sigma(\cdot) \) is the activation function applied element-wise.

More recently, several successful variations of the spectral method have been proposed, e.g., S-GCN~\cite{wu2019simplifying} and later SIGN~\cite{rossi2020sign}. S-GCN is a simplified GCN model but does not come with any performance compromises in terms of graph summarization. The idea behind the S-GCN model is to first convert large convolutional filters into smaller ones and then remove the final non-linear layers. Inspired by previous ConvGNN models, Rossi et al.~\cite{rossi2020sign} subsequently proposed SIGN, which scales ConvGNNs to extremely large graphs by combining various amendable graph convolutional filters for faster training and sampling purposes.

Another prominent line of research in spectral-based ConvGNN approaches revolves around transforming graph objects, e.g., embedding. For example, Jiang et al.~\cite{jiang2020hi} introduced a hierarchical ConvGNN for graph embedding. This team built upon a spectral ConvGNN to provide an effective representation learning scheme for end-to-end graph classification tasks. More specifically, they proposed a framework for learning graph feature embeddings while also taking the network architecture and relationships between subjects into account. Deng et al.~\cite{deng2019graphzoom} introduced a multilevel framework to enhance the scalability and accuracy of embedding in an unsupervised manner. The model initially generates a new, efficient graph that contains information about both the node properties and the topology of the original graph. It then repeatedly aggregates the nodes with high spectral similarity, breaking the graph down into numerous smaller graphs. 

\subsubsection{Spatial-based Approaches}

Spatial-based methods work on the local neighborhood of nodes, aggregating node representations from neighboring nodes to understand their properties. ConvGNNs of this kind imitate the convolution operations of CNNs by defining convolutions directly in the graph domain. Unlike spectral-based approaches, which are relatively expensive and time-consuming to compute, the structure of the spatial-based approaches is simple and has generated cutting-edge outcomes with graph summarization challenges~\cite{ajit2020review}.

As a closely-related approach to Kipf and Welling’s model~\cite{kipf2016semi}, GraphSAGE extends their framework to the inductive setting~\cite{hamilton2017inductive}. GraphSAGE was the first approach to introduce node-wise sampling coupled with minibatch training for node embeddings using spatial convolutions. The updated propagation rule, which uses a mean aggregator function, is formulated as follows~\cite{hamilton2017inductive}:
\begin{equation}
\label{equ:hamilton2017}
h_i^{(l+1)} = \sigma(W^{(l+1)} \cdot \text{AGGREGATE}(\{h_j^{(l)}, \forall j \in \mathcal{N}_i\})) 
\end{equation}
where a mean aggregator function,\(\text{AGGREGATE}(\cdot)\), combines the features of node \( i \) and its sampled neighbors. The aggregation function can be a mean, max-pooling, or any other form of aggregation. For example, the mean aggregation can be defined as:

\begin{equation}
\label{equ:hamilton2017_1}
\text{AGGREGATE}(\{h_j^{(l)}, \forall j \in \mathcal{N}_i\}) = \frac{1}{|\mathcal{N}_i|} \sum_{j \in \mathcal{N}_i} h_j^{(l)}
\end{equation}
where \( h_j^{(l)} \) is the feature vector of node \( j \) at layer \( l \) and \( |\mathcal{N}_i| \) is the number of sampled neighbors of node \( i \).

The updated feature \( h_i^{(l+1)} \) of node \( i \) is obtained by applying a learnable weight matrix \( W^{(l+1)} \) to the aggregated feature representation:

\begin{equation}
\label{equ:hamilton2017_2}
h_i^{(l+1)} = \sigma(W^{(l+1)} \cdot \text{AGGREGATE}(\{h_j^{(l)}, \forall j \in \mathcal{N}_i\})) 
\end{equation}
where \( h_i^{(l+1)} \) is the updated feature vector of node \( i \) in the \((l+1)\)-th layer. \( W^{(l+1)} \) is the weight matrix for the \( (l+1) \)-th layer. And, \( \sigma(\cdot) \) is the activation function (e.g., ReLU) applied element-wise. The resulting vector has twice the dimensionality of the input feature vector, as it contains both the original node features and the aggregated neighbor features. GraphSAGE is a more flexible model that allows for different types of aggregator functions to be used. This makes it a good choice for graphs with heterogeneous node types, where different types of information need to be aggregated differently. GraphSAGE has also been shown to perform well on tasks such as graph classification and node classification, both of which are closely tied to the task of graph summarization.


Chen et al.~\cite{chen2018fastgcn} introduced FASTGCN, a node-wise sampling approach that utilizes importance sampling for better summaries. By sampling only a fraction of nodes and leveraging importance weights, FASTGCN approximates the graph convolution operation while maintaining a high level of performance. This results in faster convergence during training, making it particularly suitable for large-scale graph-based tasks. Later, Huang et al.~\cite{huang2018adaptive} and Zeng et al.~\cite{zeng2019graphsaint} proposed layer-wise and graph-wise sampling methods, respectively, to further improve performance. Huang et al.~\cite{huang2018adaptive} focused on addressing redundancy in node-wise sampling, while Zeng et al.'s GraphSAINT~\cite{zeng2019graphsaint} aimed to correct bias and variance in minibatch estimators when sampling subgraphs for training. GraphSAINT is particularly useful for graph-based tasks where dealing with large-scale graphs can be computationally challenging. Its sampling-based approach and minibatch correction mechanism make it a powerful tool for scalable and accurate graph summarization. In a recent variation, Li et al.~\cite{li2020hierarchical} introduced bipartite GraphSAGE, tailored for bipartite graphs containing different node types and inter-layer edges. This framework involves a user-item graph supporting both user and item embedding, with nodes embedded into two separate feature spaces — one for user-related information and the other for item-related information.

Many aggregation-based methods have also been introduced to summarize graphs without sacrificing too much information within the original graphs. The summarized graph can then be used to assist in further network analysis and graph mining tasks. For instance, Yan et al.~\cite{yan2019groupinn} introduced a novel approach called GroupINN, which enhances ConvGNN through summarization, leading to faster training, data denoising, and improved interpretability. They employed an end-to-end neural network model with a specialized node grouping layer, which effectively summarizes the graph by reducing its dimensionality. Hu et al.~\cite{hu2019hierarchical} took structural similarity into consideration, aggregating nodes that share a similar structure into hypernodes. The summarized graph is then refined to restore each node's representation. To this end, a deep hierarchical ConvGNN (H-GCN) architecture with several coarsening operation layers followed by several refinement operation layers performs semi-supervised node classification. The refinement layers are designed to restore the structure of the original graph for graph classification tasks.

The most recent developments in ConvGNNs demonstrate the exciting potential graph summarization holds for a range of applications in healthcare and human motion analysis~\cite{wen2022mvs,dang2021msr}. Wen et al.~\cite{wen2022mvs}, for example, presented a promising approach to diagnosing autism spectrum disorder by parsing brain structure information through a multi-view graph convolution network. Dang et al.~\cite{dang2021msr} introduced a new type of graph convolution network, called a multi-scale residual graph convolution network, that shows superior performance in predicting human motion compared to other state-of-the-art models.  


\begin{table*}[t]
\centering
\caption{Comparative analysis of GAE-based approaches for Graph Summarization.}
\label{tab:gae_comparison}
\begin{adjustbox}{max width=\textwidth}
\begin{tabular}{m{2cm}m{1cm}m{2.5cm}m{2cm}m{1.5cm}m{4cm}m{4cm}}
\toprule
\textbf{Ref.} & \textbf{Model Name} & \textbf{Evaluation} & \textbf{Performance Metrics} & \textbf{Training Data} & \textbf{Advantages} & \textbf{Limitations} \\
\midrule
Kipf et al.~\cite{kipf2016variational} & VGAE & Link Prediction & AP, AUC & Labeled, Unlabeled & Generative model, unsupervised learning, scalable to large graphs, variational inference. & Limited expressive power, dependency on the quality and quantity of the training data, graph structure assumptions.\\
Wang et al.~\cite{wang2017mgae} & MGAE & Graph Clustering & Accuracy, F1-score, Precision, Recall, ARI & Labeled & Integration of structure and content information, deep marginalized architecture, efficient training procedure. & Performance on recall metric, limited improvement over other algorithms, static setting for structure and content.\\
Hajiramezanali et al.~\cite{hajiramezanali2019variational} & VGRNN & Link Prediction & AP, AUC & Labeled, Unlabeled & Flexibility in modeling, interpretable latent representation, incorporation of stochastic latent variables. & Marginal improvement with more flexible priors, challenges in predicting very sparse graphs, tractability of direct optimization.\\
Fan et al.~\cite{fan2020one2multi} & One2Multi & Graph Clustering & Accuracy, AUC, NMI, ARI & Labeled, Unlabeled & Effective fusion of multi-view information, joint optimization of embedding and clustering. & Time-consuming, introduction of noise, limited capacity for deep relations.\\
Cai et al.~\cite{cai2021grae} & GRAE & Graph Clustering & Accuracy, AUC, NMI, ARI & Labeled, Unlabeled & Effective fusion of multiple views, adaptive weight learning, self-training clustering. & Parameter sensitivity, dataset dependency, assumption of homogeneous graphs. \\
Salha et al.~\cite{salha2021simple} & Linear AE & Link Prediction, Node Clustering &  Average Precision, AUC-ROC, AMI & Labeled, Unlabeled & Analysis on 17 real-world graphs with various sizes and characteristics, one-hop interactions. & Performance variation across datasets, relevance of benchmark datasets, limited evaluation of deeper models.\\
Mrabah et al.~\cite{mrabah2022rethinking} & R-GAE & Graph Clustering & Accuracy, NMI, ARI & Labeled & Improved clustering performance, controlled trade-off between FR and FD, theoretical and empirical support, organized approach, and flexibility in integration. & Cannot incorporate structural and content information, trade-off between fr and fd, not suitable for generating graph-specific outputs.\\

\bottomrule
\end{tabular}
\end{adjustbox}
\end{table*}

\subsection{GAE-based Approaches} 
An autoencoder is a neural network that consists of an encoder and a decoder. Generally, the encoder transforms the input data into a condensed representation, while the decoder reconstructs the actual input data from the encoder's output~\cite{pinaya2020autoencoders}. Graph autoencoders, or GAEs, are a type of GNN that can be applied over graph structures, allowing the model to learn a compact and informative representation of a graph. Lately, GAEs have garnered increasing interest for their ability to summarize graphs due to their significant potential for dimensionality reduction~\cite{cai2018comprehensive}.

The structure of the encoder and decoder in a GAE can vary depending on the specific implementation and the complexity of the graph data. Generally, both the encoder and decoder are neural network architectures that are designed to process graph data efficiently~\cite{zhang2020deep}. 
The architecture of the encoder may include multiple layers of graph convolutions or aggregations, followed by non-linear activation functions. The output of the encoder is a compact and informative representation of the graph in the latent space. On the other hand, the decoder takes the latent representation obtained from the encoder and reconstructs the original graph structure from it. The decoder's architecture should mirror the encoder's architecture in reverse. It transforms the latent representation back into a graph structure~\cite{zhou2020graph}. 

The goal of a GAE is to learn an encoder and decoder that reduces the gap between the original graph and the reconstruction error of the decoded graph, while also encouraging the latent representation to capture meaningful information about the graph’s structure~\cite{hou2022graphmae}: 

\begin{equation}
    \label{equ:GAE}
    Z = f_E(A,X), G' = f_D(A,Z) 
\end{equation}
where \( G' \) represents the reconstructed graph, which can consist of either reconstructed features, graph structure, or both. \( A \) is the adjacency matrix, and \( X \) is the input node feature matrix. \( f_E \) serves as the graph encoder, responsible for transforming the graph and node features into a condensed representation. Conversely, \( f_D \) acts as the graph decoder, responsible for reconstructing the original graph or its components from the latent representation.

GAEs can be trained using various loss functions, such as mean squared error (MSE) or binary cross-entropy (BCE). They can also be extended to incorporate additional constraints or regularization techniques to improve the quality of the learned graph representation~\cite{salha2019keep}. For example, for graph reconstruction, the goal is to minimize the difference between the original adjacency matrix \( A \) and the reconstructed adjacency matrix \( \hat{A} \). The MSE loss is calculated as follows~\cite{kipf2016variational}:
\begin{equation}
    \label{equ:GAE_loss1}
    L_{MSE} = \frac{1}{N \times N} \sum_{i,j} (A_{ij} - \hat{A}_{ij})^2 
\end{equation}

where, \( N \) is the number of nodes in the graph, and \( A_{ij} \) and \( \hat{A}_{ij} \) are the elements of the original and reconstructed adjacency matrices, respectively.

The majority of GAE-based approaches for graph summarization use combined architectures that include ConvGNNs or RecGNNs~\cite{kipf2016variational,kingma2013auto,hajiramezanali2019variational}. For example, Kipf et al.~\cite{kipf2016variational} proposed a variational graph autoencoder (VGAE) for undirected graphs based on their previous work on spectral convolutions~\cite{kipf2016semi}. VGAE incorporates a two-layer ConvGNN model based on the variational autoencoder in~\cite{kingma2013auto}. The main concept of VGAE is to represent the input graph data not as a single point in the latent space but as a probability distribution. This distribution captures the uncertainty and variability in the graph's latent representation. Instead of directly obtaining a fixed latent representation from the encoder, VGAE samples a random point from the learned distribution. The encoder in VGAE typically consists of two or more graph convolutional layers that process the input graph data and produce latent node representations. Each graph convolutional layer can be defined as follows~\cite{kipf2016variational}:
\begin{equation}
    \label{equ:VGAE}
     Z^{(l+1)} = \sigma(\widetilde{D}^{-1/2} \widetilde{A} \widetilde{D}^{-1/2} Z^{(l)} W^{(l)})
\end{equation}
where \( Z^{(l)} \) represents the latent node representations at layer \( l \) of the encoder. \( \widetilde{A} \) is the adjacency matrix of the graph with added self-loops. \( \widetilde{D} \) is the diagonal degree matrix of \( \widetilde{A} \). \( W^{(l)} \) is the weight matrix for the \( l \)th layer. \( \sigma(\cdot) \) is the activation function (e.g., ReLU) applied element-wise.

The VGAE introduces stochasticity to GAEs by sampling the latent representation \( Z \) from a Gaussian distribution in the latent space. The mean \( \mu \) and log-variance \( \log \sigma^2 \) of the distribution are obtained from the output of the last graph convolutional layer:
\begin{equation}
    \label{equ:VGAE_1}
    \mu = Z^{(L)} \cdot W^{(\mu)}
\end{equation}
\begin{equation}
    \label{equ:VGAE_2}
    \log \sigma^2 = Z^{(L)} \cdot W^{(\sigma)}
\end{equation}

Here, \( L \) represents the last layer of the encoder. \( W^{(\mu)} \) and \( W^{(\sigma)} \) are learnable weight matrices for obtaining the mean and log-variance, respectively.

To sample from the Gaussian distribution, the reparameterization trick~\cite{kingma2015variational} is used. A random noise vector \( \epsilon \) is sampled from a standard Gaussian distribution (\( \epsilon \sim \mathcal{N}(0, 1) \)). The sampled latent representation \( Z \) is then computed as:
\begin{equation}
    \label{equ:VGAE_3}
    Z = \mu + \epsilon \cdot \exp(\frac{1}{2} \log \sigma^2)
\end{equation}

Finally, the decoder maps the sampled latent representation \( Z \) back into the graph space. In VGAE, the reconstruction is typically performed using an inner product between the latent node representations to predict the adjacency matrix \( \hat{A} \):
\begin{equation}
    \label{equ:VGAE_4}
    \hat{A} = \sigma(Z \cdot Z^T) 
\end{equation}

Here, \( \sigma(\cdot) \) is the sigmoid activation function to ensure that the predicted adjacency matrix \( \hat{A} \) is within the range [0, 1].

The loss function in VGAE consists of two terms: a reconstruction loss and a kullback-leibler (KL) divergence loss~\cite{kim2021comparing}. The reconstruction loss measures the difference between the predicted adjacency matrix \( \hat{A} \) and the actual adjacency matrix \( A \). The KL divergence loss penalizes the deviation of the learned latent distribution from the standard Gaussian distribution. The overall loss function is the sum of these two losses as follows:
\begin{equation}
    \label{equ:VGAE_l}
     L = \E_{q(Z|X,A)} [logp(A|Z)] - KL[a(Z|X,A) || p(Z)]
\end{equation}

As an extension to VGAE,  Hajiramezanali et al.~\cite{hajiramezanali2019variational} constructed a variational graph RNN by integrating a RecGNN and a GAE to model the dynamics of the node attributes and the topological dependencies. The aim of the model is to learn an interpretable latent graph representation as well as to model sparse dynamic graphs.

There are also several aggregation-based approaches built on GAEs. These are generally designed to formulate the challenges with graph clustering tasks as a summarization problem~\cite{cai2021grae,wang2017mgae,fan2020one2multi,mrabah2022rethinking}. For example, Cai et al.~\cite{cai2021grae} suggested a graph recurrent autoencoder model for use in clustering attributed multi-view graphs. The fundamental concept behind the approach is to consider both the characteristics that all views have in common and those that make each graph view special. To this end, the framework includes two separate models: the Global Graph Autoencoder (GGAE) and the Partial Graph Autoencoder (PGAE). The purpose of the GGAE is to learn the characteristics shared by all views, while the PGAE captures the distinct features. The cells are grouped into clusters using a soft K-means clustering algorithm after the output is obtained. Fan et al.~\cite{fan2020one2multi} introduced the One2Multi Graph Autoencoder (OMGAE) for multi-view graph clustering. OMGAE leverages a shared encoder to learn a common representation from multiple views of a graph and uses multiple decoders to reconstruct each view separately. Additionally, OMGAE introduces a new attention mechanism that assigns different weights to each view during the clustering process based on their importance. The model is trained to minimize a joint loss function that considers both the reconstruction error and the clustering performance. Mrabah et al.~\cite{mrabah2022rethinking} devised a new graph autoencoder model for attributed graph clustering called GAE-KL. The model uses a new formulation of the objective function, which includes a KL-divergence term, to learn a disentangled representation of the graph structure and the node attributes. The disentangled representation is then used to cluster the nodes based on their similarity in terms of both structure and attributes. The authors also introduced a new evaluation metric called cluster-based classification accuracy (CCA) to measure clustering performance. 

Recently, Salha et al.~\cite{salha2021simple} proposed a graph autoencoder architecture that uses one-hop linear models to encode and decode graph information. The approach simplifies the model while still achieving high performance with graph summarization tasks, such as node clustering and graph classification. Uniquely, this paper presents a direction for designing graph autoencoder models that balances performance with simplicity.


\begin{table*}[t]
\centering
\caption{Comparative analysis of GAT-based approaches for Graph Summarization.}
\label{tab:gat_comparison}
\begin{adjustbox}{max width=\textwidth}
\begin{tabular}{m{2cm}m{1cm}m{2.5cm}m{2cm}m{1.5cm}m{4cm}m{4cm}}
\toprule
\textbf{Ref.} & \textbf{Model Name} & \textbf{Evaluation} & \textbf{Performance Metrics} & \textbf{Training Data} & \textbf{Advantages} & \textbf{Limitations} \\
\midrule
Velivckovic et al.~\cite{velivckovic2017graph} & GAT & Node Classification & Accuracy, Micro F1-score & Labeled, Unlabeled & Adaptive attention mechanism for focusing on relevant nodes, the ability to capture long-range dependencies in graphs, scalability to handle large graph structures. & Computational complexity for large graphs, memory-intensive training, susceptible to over-smoothing.\\
Xie et al.~\cite{xie2020mgat} & MGAT & Node Classification, Link Prediction & Micro F1-score, AUC & Labeled & End-to-end multi-view graph embedding framework, attention-based integration of node information, effective and efficient performance, and generalizability to complex graph networks. & Lack of consideration for temporal dynamics, limited scalability to complex graph networks, overfitting risk with large regularization terms, lack of comparison with other multi-view embedding methods.\\
Salehi et al~\cite{salehi2020graph} & GATE & Transductive and Inductive Node Classification & Accuracy & Labeled & Inductive learning capability, a flexible and unified architecture, efficiency and scalability, and comprehensive quantitative and qualitative evaluation.& Dependency on graph structure, lack of label information, and the need for unified architectures for both transductive and inductive tasks.\\
Brody et al~\cite{brody2021attentive} & GAT v2 & Node Classification, Link Prediction & Accuracy, ROC-AUC & Labeled, Unlabeled & Addresses the limitations of the GAT model, more robust to noise, can handle more complex interactions between nodes, improved accuracy. &  Problem and dataset dependence, difficulty in predicting best architecture, performance gap between theoretical and practical models.\\
Tu et al~\cite{tu2021conditional} & KCAN & Recommender Systems & Hit@k, NDCG@k, AUC & Labeled & Effective knowledge graph distillation,  knowledge graph refinement, significant improvements, preserving local preference. & Sampling bias, time complexity, explainability, scalability, hyperparameter sensitivity. \\
Chen et al~\cite{chen2022multi} & MV-GAN & Recommender Systems & Recall@k, NDCG@k & Labeled, Unlabeled & Leveraging to address data sparsity, multi-view graph embedding, view-level attention mechanism, interpretability. & Limited consideration of factors, complexity and scalability, limited consideration of multiple modalities. \\
Li et al~\cite{li2022multi} & MRGAT & Graph Classification, Node Classification, Link Prediction & MR, MRR, Hits@k & Labeled, Unlabeled & Selective aggregation of informative features, effective fusion of entity and relation features, interpretability, and consideration of computational efficiency. & Complexity and redundant computation, sampling useful incorrect training examples,  exploiting other background knowledge, influence of attention head.\\

\bottomrule
\end{tabular}
\end{adjustbox}
\end{table*}

\subsection{GAT-based Approaches} 

The idea of an attention mechanism was first proposed by Bahdanau and his colleagues in 2014~\cite{bahdanau2014neural}. The goal was to allow for modelling long-term dependencies in sequential data and to improve the performance of autoencoders. Essentially, attention allows the decoder to concentrate on the most relevant part of the input sequence with the most relevant vectors receiving the highest weights. Graph attention networks or GATs~\cite{velivckovic2017graph} are based on the same idea. They use attention-based neighborhood aggregation, assigning different weights to the nodes in a neighborhood. This type of model is one of the most popular GNN models for node aggregation, largely because it reduces storage complexity along with the number of nodes and edges. The key formulation for a GAT is:
\begin{equation}
\label{eqn:GAT}
h_i = \sigma(\sum_{i\in N(j)} \alpha_{ij} W  h_j)
\end{equation}
where $h_i$ is the hidden feature vector of node $u_i$, $N(j)$ is the set of neighbouring nodes of $u_i$, $h_j$ is the hidden state of neighbouring node $u_j$, $W$ is a weight matrix, and $\alpha_{ij}$ is the attention coefficient that measures the importance of node $u_j$ to node $u_i$.The attention coefficients are computed as:
\begin{equation}
\label{eqn:GAT_1}
\alpha_{ij} = softmax_j(e_{ij}) = \frac{exp(e_{ij})}{\sum_{k\in N_i}exp(e_{ik})} 
\end{equation}
where $e_{ij}$ is a scalar energy value computed as:
\begin{equation}
\label{eqn:GAT_2}
 e_{ij} = LeakyReLU(a^T .[W h_i || W h_j]) 
\end{equation}
where $a$ is a learnable parameter vector, and $||$ denotes concatenation. The $LeakyReLU$ function introduces non-linearity into the model and helps prevent vanishing gradients. The softmax function normalizes the energy values across all neighboring nodes so that the attention coefficients sum to one.

By computing attention coefficients for neighboring nodes, GATs are able to selectively focus on the most important parts of the graph for each node. This allows the model to adaptively adjust to different graph structures and tasks. The attention mechanism also means GATs can incorporate node and edge features into the model, making them well-suited to summarization tasks, such as node classification with complex graphs~\cite{velickovic2017graph}.

Today, GATs are considered to be one of the most advanced models for learning with large-scale graphs. However, recently Brody et al.~\cite{brody2021attentive} argued that GATs do not actually compute dynamic attention; rather, they only compute a restricted form of static attention. To support their claim, they introduced GATv2, a new version of this type of attention network, which is capable of expressing problems that require computing dynamic attention. Focusing on the importance of dynamic weights, these authors argue that the problem of only supporting static attention can be fixed by changing the sequence of internal processes in the GAT equations, as shown in Equation \ref{equ:brody2021}. 
\begin{align}
\label{equ:brody2021}
e_{ij} = a^T LeakyReLU( W . [h_i||h_j]) 
\end{align}

As another variation of GAT, Xie et al.~\cite{xie2020mgat} proposed a novel multi-view graph attention network named MGAT,  to support low-dimensional representation learning based on an attention mechanism in a multi-view manner. The authors focus on a view-based attention approach that not only aggregates view-based node representations but also integrates various types of relationships into multiple views. 

Tu et al.~\cite{tu2021conditional} explored the benefits of using graph summarization and refining bipartite user-item graphs for recommendation tasks. They applied a conditional attention mechanism to task-based sub-graphs to determine user preferences, which emphasizes the potential of summarizing and enhancing knowledge graphs to support recommender systems. Salehi et al.~\cite{salehi2020graph} defined a model based on an autoencoder architecture with a graph attention mechanism that learns low-dimensional representations of graphs. The model compresses the information in the input graph into a fixed-size latent vector, which serves as a summary of the entire graph. Through the use of attention, the model is able to discern and prioritize critical nodes and edges within the graph, making it more effective at capturing the graph's structural and semantic properties.

More recent works on GATs conducted by Chen et al.~\cite{chen2022multi} and Li et al.~\cite{li2022multi} demonstrate the potential of graph attention networks for summarizing and analyzing complex graph data in various domains. Chen et al. proposed a multi-view graph attention network for travel recommendations. The model takes several different types of user behaviors into account, such as making hotel reservations, booking flights, and leaving restaurant reviews, and, in the process, learns an attention mechanism to weigh the importance of different views for a recommendation. Li et al. developed a multi-relational graph attention network for knowledge graph completion. The model integrates an attention mechanism and edge-type embeddings to capture the complex semantic relations between entities in a knowledge graph. 


\section{Graph Reinforcement Learning}
\label{GRL}


Reinforcement learning (RL) is a mathematical model based on sequential decisions that allows an agent to learn via trial and error in an interactive setting through feedback on its actions. Due to the success and fast growth of reinforcement learning in interdisciplinary fields, scholars have recently been inspired to investigate reinforcement learning models for graph-structured data, i.e., graph reinforcement learning or GRL~\cite{mingshuo2022reinforcement}. GRL is largely implemented based on the Bellman theory~\cite{bellman1964selected}, where the environment is represented as a graph, nodes represent states, edges represent possible transitions between states, and rewards are associated with specific state-action pairs or nodes. The key components of GRL are as follows~\cite{mingshuo2022reinforcement}:

\begin{itemize}
    \item Environment ($graph$): The graph $G$ represents the environment in which the agent operates. It is defined as $G = (V, E)$, where $V$ is the set of nodes representing states and E is the set of edges representing possible transitions between states.
    \item State ($s$): In GRL, a state $s$ corresponds to a specific node in the graph. Each node may have associated attributes or features that provide information about the state.
    \item Action ($a$): An action $a$ corresponds to a decision or move that the agent can make when in a particular state (node). In graph-based environments, actions can be related to traversing edges between nodes or performing some operation on a node.
    \item Transition Model ($T$): The transition model defines the dynamics of the graph, specifying the probability of moving from one state (node) to another by taking a specific action (edge).\\
    $T(s, a, s') = P(s' | s, a)$, where $s$ is the current state (node), $a$ is the action (edge), and $s'$ is the next state (node).
   \item Reward Function ($R$): The reward function defines the immediate reward the agent receives after taking a particular action in a given state (node). \\
   $R(s, a) =$ Expected immediate reward received when taking action a in state $s$.
   \item Policy ($\pi$): Similar to standard RL, the policy in Graph RL is a strategy that the agent uses to decide which action to take in each state (node).\\ $\pi(a | s) = $  Probability of taking action $a$ in state $s$.
   \item Value Function ($V$) and Q-function ($Q$): In Graph RL, the value function V(s) and the Q-function $Q(s, a)$ represent the expected cumulative reward the agent can obtain starting from a particular state (node) and following a policy $\pi$, or by taking action $a$ in state $s$ and then following policy $\pi$, respectively. The Q-learning algorithm can be formulated as:
   \begin{multline}
       \label{equ:qlearning}
       Q(s_t,a_t)\leftarrow  Q(s_t,a_t)+\alpha[R_{t+1} +\\ {\gamma max_a  Q(s_{t+1},a)- Q(s_t,a_t)]}
   \end{multline}
    
    where at each timestep \(t\), the state \(s_t\) interacts with the environment using a behavior policy based on the Q-table values. It takes action \(a\), receives a reward \(R\), and transitions to a new state \(s_{t+1}\) based on the environment's feedback. This process is used to update the Q-table iteratively, continually incorporating information from the new state \(s_{t+1}\) until reaching the termination time \(t\).
    \end{itemize}

The primary objective in GRL is to acquire a policy that maximizes the expected Q-function $Q(s,a$) over a sequence of actions, the target policy is defined as~\cite{nie2023reinforcement}:
\begin{multline}
    \label{equ:grl}
    \pi^* = \arg\max_\pi Q(s,a) \\
    = \arg\max_\pi \E_{\pi,T}[\sum_{k=0} \gamma^k r_{t+k} | s_t = s, a_t = a]
\end{multline}
where \(\E_{\pi, T}[\cdot]\) denotes the expectation with respect to both the policy \(\pi\) and the distribution of transitions \(T\) (i.e., state transitions and rewards). The expression \(\sum_{k=0} \gamma^k r_{t+k}\) represents the sum of discounted rewards obtained in the future starting from time step \(t\) (the current time step) and continuing for \(k\) steps into the future. \(r_{t+k}\) is the reward obtained at time step \(t+k\) after taking action \(a_t\) at time step \(t\) and following policy \(\pi\) thereafter. The discount factor \(\gamma\) is a value between 0 and 1 that determines the importance of immediate rewards compared to future rewards. It discounts future rewards to make them less significant than immediate rewards. Smaller \(\gamma\) values make the agent more myopic, whereas larger \(\gamma\) values make the agent more far-sighted. The overall objective is to find the policy \(\pi\) that maximizes the expected sum of rewards (the Q-function) starting from state \(s\) and taking action \(a\). 

Achieving this goal involves employing various algorithms, like Q-learning, or utilizing a policy gradient method that updates Q-values or policy parameters based on observed rewards and transitions~\cite{zhang2020deep}. 

GRL employs a diverse range of algorithms, and it frequently utilizes GNNs to efficiently process and learn from data structured as graphs. GNNs play a crucial role in updating node representations by considering their neighboring nodes, and they are seamlessly integrated into the RL framework to handle tasks specific to graphs with effectiveness. They are seamlessly integrated into the graph summarization framework to effectively handle tasks that involve summarizing graph structures. For instance, Yan et al.~\cite{yan2020Automaticgrl} introduced a ConGNN-based neural network specifically designed for graph sampling, enabling the automatic extraction of spatial features from the irregular graph topology of the substrate network. To optimize the learning agent, they adopt a popular parallel policy gradient training method, enhancing efficiency and robustness during training. 
Wu et al.~\cite{Wu2020signalgrl} tackled the problem of graph signal sampling by formulating it as a reinforcement learning task in a discrete action space. They use a deep Q-network (DQN) to enable the agent to learn an effective sampling strategy. To make the training process more adaptable, they modify the steps and episodes. During each episode, the agent learns how to choose a node at each step and selects the best node at the end of the episode. They also redefine the actions and rewards to suit the sampling problem. In another work by Wu et al.~\cite{Wu2022Reinforced}, a reinforced sample selection approach for GNNs' transfer learning is proposed. The approach uses GRL to guide transfer learning and reduce the divergence between the source and target domains. 

There is also a line of GRL research that seeks to use this paradigm to evaluate and improve the quality of graph summaries. For example, Amiri et al.~\cite{amiri2018netgist} introduced a task-based GRL framework to automatically learn how to generate a summary of a given network. To provide an optimal solution for finding the best task-based summary, the authors made use of CNN layers in combination with a reinforcement learning technique. To improve the quality of the summary, the authors later proposed NetReAct~\cite{amiri2020netreact}, an interactive learning framework for graph summarization. The model uses human feedback in tandem with reinforcement learning to improve the summaries, while visualizing the document network. 

\begin{table*}[t]
\centering
\caption{Comparative analysis of selected GRL-based approaches for Graph Summarization.}
\label{tab:gRL_comparison}
\begin{adjustbox}{max width=\textwidth}
\begin{tabular}{m{2cm}m{1cm}m{2.5cm}m{2cm}m{1.5cm}m{4cm}m{4cm}}
\toprule
\textbf{Ref.} & \textbf{Model Name} & \textbf{Evaluation} & \textbf{Performance Metrics} & \textbf{Training Data} & \textbf{Advantages} & \textbf{Limitations} \\
\midrule
Wu et al.~\cite{Wu2020signalgrl} & DQN & Graph Reconstruction & Accuracy & Labeled, Unlabeled & Efficient graph sampling, RL approach, no need for labeled data, potential for automation, improved reconstruction accuracy. & Limitations on performance on big graphs, sampling set size, the assumptions on the training graph.\\
Amiri et al.~\cite{amiri2018netgist} & NetGist & Influence Maximization, Community Detection & $\rho_{netgist}$ & Labeled, Unlabeled  & Automatic flexible approach to generating meaningful graph summaries for a given set of tasks, generalization to unseen Instances. & Computational complexity, task dependency, limited scope of graph optimization problems, lack of comparison with existing methods.\\
Amiri et al.~\cite{amiri2020netreact} & NetReAct & Graph Clustering & $\rho_{netreact}$ & Labeled, Unlabeled & Incorporating human feedback, meaningful relationships between groups, multi-scale visualization & The simplicity of non-expert feedback, sparsity and inconsistency of human feedback, scalability to larger document datasets, need for further exploration and development.\\
Wickman et al.~\cite{wickman2021sparrl} & SparRL & PageRank, community structure & Spearman’s $\rho$ index, ARI & Labeled, Unlabeled & Task adaptability, learning efficiency and convergence, flexibility, efficiency, and ease of use. & Involves matrix operations and can be computationally intensive for large graphs, sampling-based techniques, limited to static graph settings.\\
Wu et al.~\cite{Wu2022Reinforced} & RSS-GNN & Transfer Learning & AUC-ROC & Labeled, Unlabeled & Efficient and effective sample selection, alleviates divergence between source and target domain graphs. & Non-differentiable sample selection,  computational complexity, generalization to new downstream tasks.\\
Zhao et al.~\cite{zhao2022deep} & BNN-GNN & Graph Classification & Average Accuracy, AUC & Labeled, Unlabeled & Improved classification performance, customized aggregation, effective brain network analysis, flexibility in meta-policy application, robustness to different input types. & Performance variation with input types, generalizability, hyperparameter sensitivity, explainability. \\
Goyal et al.~\cite{goyal2022graph} & GNRL & Graph Classification & Topk Accuracy & Labeled, Unlabeled & Improved image representation, interpretability, efficient training, and scalability. & Lack of robustness in coarsened graph representation, training time and gpu utilization, limited improvement over model-free techniques, lack of generalizability to other environments.\\
\bottomrule
\end{tabular}
\end{adjustbox}
\end{table*}

In another study, Wickman et al.~\cite{wickman2021sparrl} recently presented a graph sparsification framework, SparRL, empowered by a GRL to be used for any edge sparsification assignment with a specific target for reduction. The model takes an edge reduction ratio as its input, and a learning model decides how best to prune the edges. 
SparRL proceeds in a sequential manner, removing edges from the graph until a total of edges have been pruned. 
In another work, Wu et al.~\cite{wu2020graph} introduced GSGAN, a novel method for graph sparsification in community detection tasks. GSGAN excels at identifying crucial relationships not apparent in the original graph and enhances community detection effectiveness by introducing artificial edges. Employing a generative adversarial network (GAN) model, GSGAN generates random walks that effectively capture the network's underlying structure. What sets this approach apart is its utilization of reinforcement learning, which enables the method to optimize learning objectives by deriving rewards from a specially designed reward function. This reinforcement learning component guides the generator to create highly informative random walks, ultimately leading to improved performance in community detection tasks. Yan et al.~\cite{yan2019spatially} introduced a GRL approach to summarize geographic knowledge graphs. To obtain a more thorough understanding of the summarizing process, the model exploits components with spatial specificity and includes both the extrinsic and the intrinsic information in the graph. The authors also discuss the effectiveness of spatial-based models and compare the results of their model with models that include non-spatial entities. 

Recently, many articles have discussed the potential of using GNN-based GRLs to summarize and analyze complex graph data in domains like neuroscience and computer vision~\cite{zhao2022deep,goyal2022graph}. For example, Zhao et al.~\cite{zhao2022deep} suggested a deep reinforcement learning scheme guided by a GNN as a way to analyze brain networks. The model uses a reinforcement learning framework to learn a policy for selecting the most informative nodes in the network and combines that with a GNN to learn the node representations. Also, Goyal et al.~\cite{goyal2022graph} presented a GNN-based approach to image classification that relies on reinforcement learning. The model represents images as graphs and learns graph convolutional filters to extract features from the graph representation. They showed that their model outperforms several state-of-the-art methods on benchmark datasets with both image classification and reinforcement learning tasks. 

In Table~\ref{tab:gRL_comparison}, we summarize the key features of representative GRL-based approaches for graph summarization. Evaluation methods, performance metrics, training data, advantages, and limitations are compared among different models. 


\begin{table*}[t]
\centering
\begin{center}
    \caption{Published Datasets.}
        \begin{tabular}{p{3cm} p{2cm} p{3.5cm} p{6cm}}
        \hline
        \textbf{Category} & \textbf{Dataset} & \textbf{Publications} & \textbf{URL} \\
        \hline
            \multirow{10}{*}{Citation Networks} &
                                   \multicolumn{1}{m{2cm}}{Cora} & \multicolumn{1}{m{3.5cm}}{\cite{kipf2016semi,kipf2016variational,velickovic2017graph,wang2017mgae,chen2018fastgcn,huang2018adaptive,wu2019simplifying,deng2019graphzoom,hu2019hierarchical,salehi2020graph,salha2021simple,ge2022gr,mrabah2022rethinking, bojchevski2018netgan}} & \multicolumn{1}{m{6cm}}{\href{https://relational.fit.cvut.cz/dataset/cora}{https://relational.fit.cvut.cz/dataset/CORA}}\\\cline{2-4}
                                 & \multicolumn{1}{m{2cm}}{Citeseer} & \multicolumn{1}{m{3.5cm}}{\cite{kipf2016semi,kipf2016variational,velickovic2017graph,wang2017mgae,huang2018adaptive,deng2019graphzoom,hu2019hierarchical,salehi2020graph,salha2021simple,ge2022gr,mrabah2022rethinking,bojchevski2018netgan}} & \multicolumn{1}{m{6cm}}{\href{https://relational.fit.cvut.cz/dataset/citeseer}{https://relational.fit.cvut.cz/dataset/CiteSeer}}\\\cline{2-4}
                                 & \multicolumn{1}{m{2cm}}{PubMed} & \multicolumn{1}{m{3.5cm}}{\cite{kipf2016semi,kipf2016variational,velickovic2017graph,huang2018adaptive,deng2019graphzoom,hu2019hierarchical,salehi2020graph,salha2021simple,ge2022gr,mrabah2022rethinking,bojchevski2018netgan}} & \multicolumn{1}{m{6cm}}{\href{https://relational.fit.cvut.cz/dataset/PubMed_Diabetes}{https://relational.fit.cvut.cz/dataset/PubMed\_Diabetes}}\\\cline{2-4}
                                 & \multicolumn{1}{m{2cm}}{DBLP} & \multicolumn{1}{m{3.5cm}}{\cite{fan2020one2multi,cai2021grae,bojchevski2018netgan, khoshraftar2019dynamic}} & \multicolumn{1}{m{6cm}}{\href{https://dblp.uni-trier.de/xml/}{https://dblp.uni-trier.de/xml/}}\\\cline{2-4}
                                 & \multicolumn{1}{m{2cm}}{ACM} & \multicolumn{1}{m{3.5cm}}{\cite{fan2020one2multi,cai2021grae,khoshraftar2019dynamic}} & \multicolumn{1}{m{6cm}}{\href{http://www.arnetminer.org/open-academic-graph}{http://www.arnetminer.org/open\-academic\-graph}}\\
            \hline
            \multirow{8}{*}{Social Networks} &
                                   \multicolumn{1}{m{2cm}}{Reddit} & \multicolumn{1}{m{3.5cm}}{\cite{hamilton2017inductive,chen2018fastgcn,huang2018adaptive,wu2019simplifying,deng2019graphzoom,zeng2019graphsaint,rossi2020sign,salha2021simple,zhang2022dynamic}} & \multicolumn{1}{m{6cm}}{\href{https://github.com/reddit-archive/reddit}{https://github.com/reddit\-archive/reddit}}\\\cline{2-4}
                                 & \multicolumn{1}{m{2cm}}{IMDB} & \multicolumn{1}{m{3.5cm}}{\cite{fan2020one2multi,cai2021grae}} & \multicolumn{1}{m{6cm}}{\href{https://datasets.imdbws.com/}{https://datasets.imdbws.com/}}\\\cline{2-4}
                                 & \multicolumn{1}{m{2cm}}{Karate} & \multicolumn{1}{m{3.5cm}}{\cite{amiri2018netgist}} & \multicolumn{1}{m{6cm}}{\href{http://networkdata.ics.uci.edu/data/karate/}{http://networkdata.ics.uci.edu/data/karate/}}\\\cline{2-4}
                                 & \multicolumn{1}{m{2cm}}{Facebook} & \multicolumn{1}{m{3.5cm}}{\cite{amiri2018netgist,hajiramezanali2019variational,wickman2021sparrl,wu2020graph}} & \multicolumn{1}{m{6cm}}{\href{http://snap.stanford.edu/data/ego-Facebook.html}{http://snap.stanford.edu/data/ego-Facebook.html}}\\\cline{2-4}
                                 & \multicolumn{1}{m{2cm}}{DNC} & \multicolumn{1}{m{3.5cm}}{\cite{ma2020streaming}} & \multicolumn{1}{m{6cm}}{\href{https://github.com/alge24/DyGNN/tree/main/Dataset}{https://github.com/alge24/DyGNN/tree/main/Dataset}}\\\cline{2-4}
                                 & \multicolumn{1}{m{2cm}}{UCI} & \multicolumn{1}{m{3.5cm}}{\cite{ma2020streaming}} & \multicolumn{1}{m{6cm}}{https://github.com/alge24/DyGNN/tree/main/Dataset}\\\cline{2-4}
                                 & \multicolumn{1}{m{2cm}}{Twitter} & \multicolumn{1}{m{3.5cm}}{\cite{xie2020mgat,wickman2021sparrl}} & \multicolumn{1}{m{6cm}}{\href{http://snap.stanford.edu/data/ego-Twitter.html}{http://snap.stanford.edu/data/ego-Twitter.html}}\\
                                 
            \hline
            \multirow{10}{*}{User-generated Networks} &
                                   \multicolumn{1}{m{2cm}}{Amazon} & \multicolumn{1}{m{3.5cm}}{\cite{zeng2019graphsaint,wickman2021sparrl}} & \multicolumn{1}{m{6cm}}{\href{http://snap.stanford.edu/data/amazon-meta.html}{http://snap.stanford.edu/data/amazon-meta.html}}\\\cline{2-4}
                                 & \multicolumn{1}{m{2cm}}{Yelp} & \multicolumn{1}{m{3.5cm}}{\cite{zeng2019graphsaint,rossi2020sign,tu2021conditional,zhang2022dynamic}} & \multicolumn{1}{m{6cm}}{\href{https://www.yelp.com/dataset}{https://www.yelp.com/dataset}}\\\cline{2-4}
                                 & \multicolumn{1}{m{2cm}}{Epinions} & \multicolumn{1}{m{3.5cm}}{\cite{ma2020streaming}} & \multicolumn{1}{m{6cm}}{\href{http://snap.stanford.edu/data/soc-Epinions1.html}{http://snap.stanford.edu/data/soc-Epinions1.html}}\\\cline{2-4}
                                 & \multicolumn{1}{m{2cm}}{Taobao} & \multicolumn{1}{m{3.5cm}}{\cite{li2020hierarchical}} & \multicolumn{1}{m{6cm}}{\href{https://tianchi.aliyun.com/dataset/649}{https://tianchi.aliyun.com/dataset/649}}\\\cline{2-4}
                                 & \multicolumn{1}{m{2cm}}{MovieLens} & \multicolumn{1}{m{3.5cm}}{\cite{tu2021conditional,chen2022multi}} & \multicolumn{1}{m{6cm}}{\href{https://grouplens.org/datasets/movielens/}{https://grouplens.org/datasets/movielens/}}\\\cline{2-4}
                                 & \multicolumn{1}{m{2cm}}{Last-FM} & \multicolumn{1}{m{3.5cm}}{\cite{tu2021conditional}} & \multicolumn{1}{m{6cm}}
                                 {\href{https://grouplens.org/datasets/hetrec-2011/}{https://grouplens.org/datasets/hetrec-2011/}}\\\cline{2-4}
                                 & \multicolumn{1}{m{2cm}}{Eumail} & \multicolumn{1}{m{3.5cm}}{\cite{wu2020graph}} & \multicolumn{1}{m{6cm}}
                                 {\href{https://snap.stanford.edu/data/email-EuAll.html}{https://snap.stanford.edu/data/email-EuAll.html}}\\\cline{2-4}
                                 & \multicolumn{1}{m{2cm}}{Enron} & \multicolumn{1}{m{3.5cm}}{\cite{hajiramezanali2019variational}} & \multicolumn{1}{m{6cm}}
                                 {\href{https://snap.stanford.edu/data/email-Enron.html}{https://snap.stanford.edu/data/email-Enron.html}}\\\cline{2-4}
                                 & \multicolumn{1}{m{2cm}}{POL. BLOGS} & \multicolumn{1}{m{3.5cm}}{\cite{bojchevski2018netgan}} & \multicolumn{1}{m{6cm}}
                                 {\href{https://networks.skewed.de/net/polblogs}{https://networks.skewed.de/net/polblogs}}\\

            \hline
            \multirow{10}{4cm}{Bio-informatic Networks, Image/Neuroimage} &
                                   \multicolumn{1}{m{2cm}}{PPI} & \multicolumn{1}{m{3.5cm}}{\cite{hamilton2017inductive,velickovic2017graph,deng2019graphzoom,zeng2019graphsaint,rossi2020sign,salha2021simple,zhang2022dynamic}} & \multicolumn{1}{m{6cm}}{\href{https://github.com/williamleif/GraphSAGE}{https://github.com/williamleif/GraphSAGE}}\\\cline{2-4}
                                 & \multicolumn{1}{m{2cm}}{MUTAG} & \multicolumn{1}{m{3.5cm}}{\cite{taheri2018learning,jin2018learning,salehi2020graph}} & \multicolumn{1}{m{6cm}}{\href{https://networkrepository.com/Mutag.php}{https://networkrepository.com/Mutag.php}}\\\cline{2-4}
                                 & \multicolumn{1}{m{2cm}}{PTC} & \multicolumn{1}{m{3.5cm}}{\cite{taheri2018learning}} & \multicolumn{1}{m{6cm}}{\href{ http://www.predictive-toxicology.org/ptc/}{ http://www.predictive-toxicology.org/ptc/}}\\\cline{2-4}
                                 & \multicolumn{1}{m{2cm}}{ENZymes} & \multicolumn{1}{m{3.5cm}}{\cite{taheri2018learning,jin2018learning}} & \multicolumn{1}{m{6cm}}{\href{https://github.com/snap-stanford/GraphRNN/tree/master/dataset/ENZYMES}{https://github.com/snap-stanford/GraphRNN/tree/master/dataset/ENZYMES}}\\\cline{2-4}
                                 & \multicolumn{1}{m{2cm}}{NCI} & \multicolumn{1}{m{3.5cm}}{\cite{taheri2018learning,jin2018learning}} & \multicolumn{1}{m{6cm}}
                                 {\href{https://cdas.cancer.gov/}{https://cdas.cancer.gov/}}\\\cline{2-4}

                                &\multicolumn{1}{m{2cm}}{Flickr} & \multicolumn{1}{m{3.5cm}}{\cite{zeng2019graphsaint,rossi2020sign}} & \multicolumn{1}{m{6cm}}{\href{https://shannon.cs.illinois.edu/DenotationGraph/}{https://shannon.cs.illinois.edu/DenotationGraph/}}\\\cline{2-4}
                                 & \multicolumn{1}{m{2cm}}{fMRI} & \multicolumn{1}{m{3.5cm}}{\cite{yan2019groupinn,zhao2022deep}} & \multicolumn{1}{m{6cm}}{\href{https://adni.loni.usc.edu/data-samples/access-data/}{https://adni.loni.usc.edu/data-samples/access-data/}}\\\cline{2-4}
                                 & \multicolumn{1}{m{2cm}}{ADNI} & \multicolumn{1}{m{3.5cm}}{\cite{jiang2020hi}} & \multicolumn{1}{m{6cm}}{\href{https://adni.loni.usc.edu/data-samples/access-data/}{https://adni.loni.usc.edu/data-samples/access-data/}}\\\cline{2-4}
                                 & \multicolumn{1}{m{2cm}}{ABIDE} & \multicolumn{1}{m{3.5cm}}{\cite{jiang2020hi}} & \multicolumn{1}{m{6cm}}{\href{https://fcon_1000.projects.nitrc.org/indi/abide/}{https://fcon_1000.projects.nitrc.org/indi/abide/}}\\\cline{2-4}
            \hline
            Knowledge Graphs & &\cite{yan2019spatially,kipf2016semi,hu2019hierarchical,yan2020Automaticgrl} & \\
            \hline
            Synthetic Networks & &\cite{amiri2018netgist,amiri2020netreact,goyal2020dyngraph2vec,brody2021attentive,zhang2022dynamic} & \\\hline

        \end{tabular}        
    \end{center}
    \label{table:datasets}
\end{table*}

\section{Published Algorithms and Datasets}
\label{resources}

In the following section, we will offer a comprehensive overview of the benchmark datasets, evaluation metrics, and open-source implementations. These critical components are extensively examined and elaborated upon in Sections~\ref{GNN} and~\ref{GRL} of the literature survey. By delving into these aspects, we aim to provide a thorough understanding of the landscape covered in the aforementioned sections.

\subsection{Datasets}
\begin{figure}[t]
  \centering
  \includegraphics[width=0.5\textwidth]{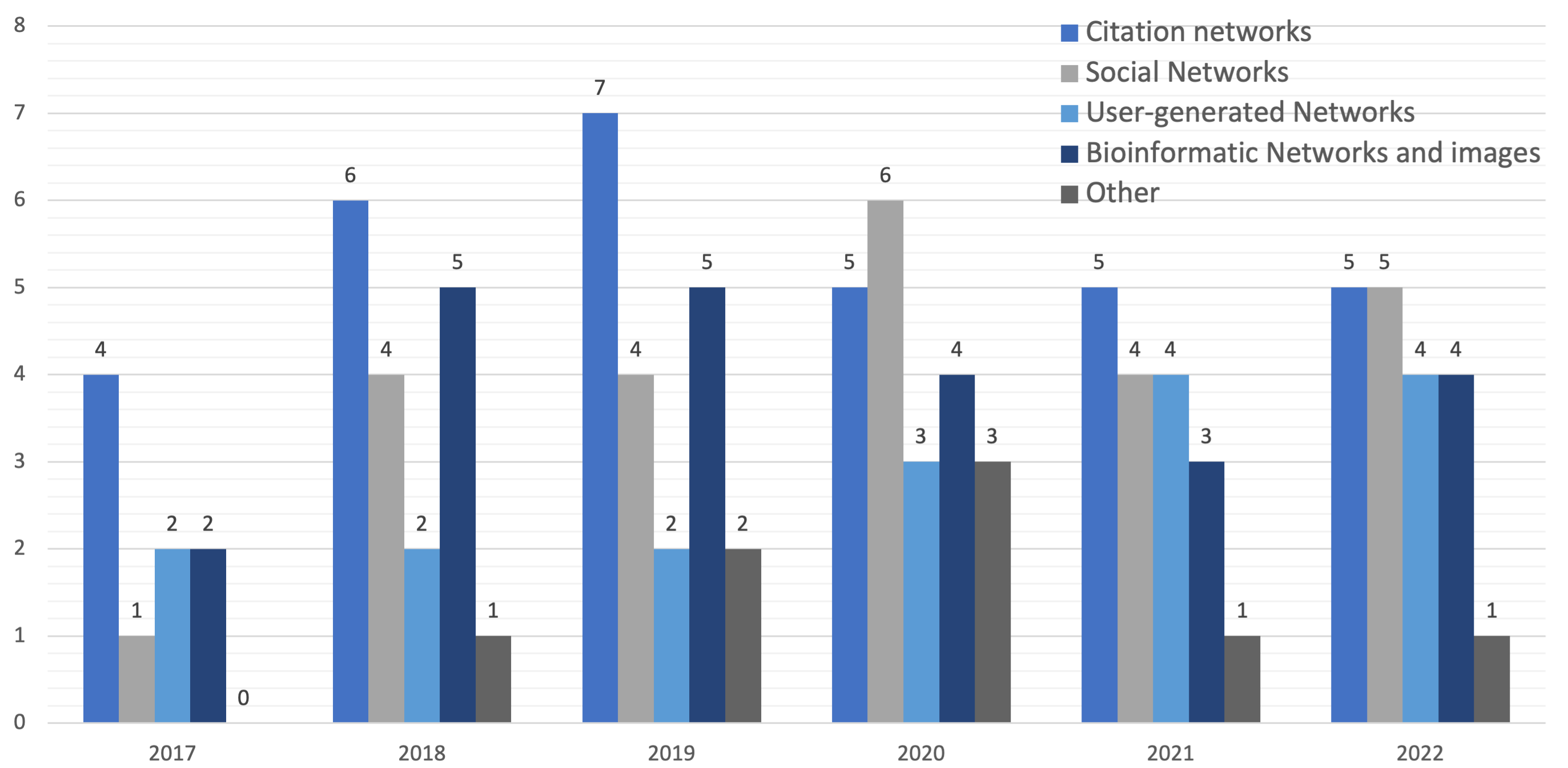}
  \caption{Year-wise development of datasets in the reviewed papers.}
  \label{fig:yw_dataset}
\end{figure}

Both synthetic and real-world datasets are used in the development of the field. Synthetic datasets are created by models based on manually-designed rules, while real-world datasets are collected from actual applications and used to evaluate the performance of proposed methods for practical use. The popular real-world datasets are divided into five categories: citation networks, social networks, user-generated networks, bio-informatics networks and image/neuroimages, and knowledge graphs. Table~\ref{table:datasets}I presents an overview of the datasets most commonly utilized in the mentioned categories. Additionally, Figure~\ref{fig:yw_dataset} offers a year-wise development trend for each, providing valuable insights into their evolution and availability over time.

\subsection{Evaluation Metrics}
GNNs are typically evaluated through tasks like node classification, graph classification, graph clustering, recommendation, and link prediction. To provide an overview of the evaluation criteria used in each study, we categorized each of the articles we reviewed based on the metrics they used to evaluate their methods. These metrics mostly include accuracy, precision, recall, F1-score, and AUC-ROC. Table~\ref{table:evaluation} lists the most-used evaluation metrics and their calculation formulas or descriptions.

\subsection{Open Source Implementations}

Tables ~\ref{table:opensource_nc} and ~\ref{table:opensource_lp} provide a detailed comparison of selected approaches, showcasing the outcomes of replicating established comparable models through two leading GNN-based graph summarization evaluation techniques: node classification and link prediction. The evaluation is conducted on both static datasets such as Cora and Citeseer, as well as dynamic datasets like Enron and Facebook to assess model performance in evolving graph structures. This comparison encompasses model names, platforms, datasets, metrics, hyperparameters, and implementation sources, with all implementations developed using Python 3.x and popular frameworks like PyTorch, PyTorch Geometric, or TensorFlow.

\section{Discussion and Future Directions}
\label{future}

This survey has provided a comprehensive examination of GNNs in the context of graph summarization, focusing on key methodologies such as RecGNN, ConvGNN, GAE, GAT, and the emerging field of GRL. Each of these approaches offers unique perspectives and methodologies for capturing the complex relationships inherent in graph data.

\textbf{RecGNN-based} approaches can capture a substantial amount of information during their recursive neighbourhood expansions by using recurrent units to identify the long-term dependence across layers.  Further, this process is quite efficient. Notably, RecGNN models can also improve numerical stability during training if they incorporate convolutional filters. However, they may face challenges in long-range dependency modeling due to the vanishing gradient problem, a common issue in recurrent architectures.

\textbf{ConvGNN-based} approaches leverage a more spatial approach, effectively aggregating local neighborhood information. This method has been particularly effective in tasks where local structure is highly informative. Nonetheless, the convolutional approach may not fully capture the global context, which can be critical in certain summarization tasks. In addition, most existing ConvGNN models for graph summarization simply presume the input graphs are static. However, in the real world, dynamically evolving graphs/networks are more common. For instance, in a social network the number of users, their connected friends, and their activities are constantly changing. To this end, learning ConvGNNs on static graphs may not yield the best results. Hence, more research on dynamic ConvGNN models is needed to increase the quality of summaries with large-scale dynamic graphs. 

\textbf{GAE-based} approaches offer a powerful framework for unsupervised learning on graphs. By learning to encode and decode graph data, GAEs can generate compact representations that preserve essential topological information. However, the quality of the summarization is heavily dependent on the choice of the encoder and decoder, which can be a non-trivial design choice. In addition, most GAE-based approaches are typically unregularized and mainly focus on minimizing the reconstruction error while ignoring the data distribution of the latent space. However, this might lead to poor graph summarization when working with sparse and noisy real-world graph data. Although there are a few studies on GAE regularization~\cite{pan2018adversarially}, more research is needed in this regard. 

\textbf{GAT-based} approaches introduce an attention mechanism that allows for the weighting of nodes' contributions to the representation. This approach can adaptively highlight important features and relationships within the graph. While GATs provide a flexible mechanism that can potentially outperform other methods, they may also require more computational resources and can be prone to overfitting on smaller datasets. Given the recent advancements in this area, we expect to see more research in the future on using GATs to create condensed representations of both static and dynamic graphs.

\begin{table}[t]
\centering
\caption{Evaluation metrics.}
\begin{tabular}{m{2.5cm}|m{5cm}}
    \hline
    \textbf{Evaluation Metric} & \textbf{Formula/Description} \\
    \hline
    Accuracy & \vspace{0.1cm} $\frac{tp + tn}{tp + tn + fp + fn}$ \\
    \hline
    Precision & \vspace{0.1cm} $\frac{tp}{tp + fp}$ \\
    \hline
    Recall & \vspace{0.1cm} $\frac{tp}{tp + fn}$ \\
    \hline
    Average Precision & \vspace{0.1cm} $\sum_{i=1}^n (R_i - R_{i-1}) \cdot P_i$ \\
    \hline
    F1-score & \vspace{0.1cm} $2 \times \frac{Recall \times Precision}{Recall + Precision}$ \\
    \hline
    Micro F1-score & \vspace{0.1cm} $\frac{tp}{tp + fp + fn}$ \\
    \hline
    Specificity & \vspace{0.1cm} $\frac{tn}{tn + fp}$ \\
    \hline
    AUC-ROC & \vspace{0.1cm} The Area Under the ROC Curve. \\
    \hline
    ARI & \vspace{0.1cm} Adjusted Rand Index, measures the similarity between two data clusterings. \\
    \hline
    NMI & \vspace{0.1cm} The Normalized Mutual Information measures the similarity between two clusters.\\
    \hline
    Hit@k & \vspace{0.1cm} $\frac{\text{Number of relevant items in top } k}{k}$ \\
    \hline
    NDCG@k & \vspace{0.1cm} Normalized Discounted Cumulative Gain at k. \\
    \hline
    Spearman's $\rho$ index & \vspace{0.1cm} Measures the strength and direction of the monotonic relationship between two variables. \\
    \hline
    $\rho_{netgist}$ & \vspace{0.1cm} Expected ratio.  \\
    \hline
    $\rho_{netreact}$ &\vspace{0.1cm}  Quantifying the ease of identifying relevant documents. \\
    \hline
    Mean Rank & \vspace{0.1cm} $ \frac{1}{n} \sum_{i=1}^{n} rank_i$ \\
    \hline
    Mean Reciprocal Rank & \vspace{0.1cm} $ \frac{1}{N} \sum_{i=1}^{N} \frac{1}{rank_i}$ \\
    \hline
\end{tabular}
\label{table:evaluation}
\end{table}

\begin{table*}[t]
    \centering
    \caption{Comparable models from published literature using Node Classification for evaluations. S/D: Static/Dynamic, NL: Number of Layers, AF: Activation Functions, DR: Dropout rate, LR: Learning Rate, WD: Weight Decay. }
    \begin{tabular}{m{0.5cm} m{2cm} m{1.5cm} m{2cm} m{1.5cm} m{1.5cm} m{5cm} m{0.5cm}}
        \hline
          & \centering \textbf{Model} & \centering \textbf{Platform} & \centering \textbf{Dataset} & \multicolumn{2}{c}{\centering \textbf{Metrics (\%)}} & \centering \textbf{Hyperparameters} & \textbf{Repo.} \\
        \hline
        \multirow{17}{*}{\rotatebox[origin=c]{90}{Node Classification}} &
          &  &  & \vspace{0.1cm} \centering \textbf{Accuracy} & \vspace{0.1cm} \centering \textbf{F1-score} &  & \\
          &  \multirow{2}{=}{GCN~\cite{kipf2016semi}} &  \centering \multirow{2}{=}{\centering PyTorch-Geometric} &  \centering Cora (S) &  \centering  81.32$\pm$0.01 & 81.11$\pm$0.01 & \multirow{2}{=}{NL, AF, DR, LR, WD, Optimizer} &  \multirow{2}{*}{\centering \href{https://pytorch-geometric.readthedocs.io/en/latest/_modules/torch_geometric/nn/conv/gcn_conv.html}{LINK}} \\
          & &  &  \centering Citeseer (S) & \centering 69.30$\pm$0.01 & 67.10$\pm$0.01 &  &  \\ 

          &  \multirow{2}{=}{GraphSAGE~\cite{hamilton2017inductive}} &  \multirow{2}{=}{\centering PyTorch-Geometric} &  \centering Cora (S) &  \centering  80.20$\pm$0.00 & 80.30$\pm$0.00 & \multirow{2}{=}{NL, AF, DR, LR, WD, Optimizer, Sampling size, Aggregator type} &  \multirow{2}{*}{\centering \href{https://pytorch-geometric.readthedocs.io/en/latest/_modules/torch_geometric/nn/conv/sage_conv.html}{LINK}} \\
          & &  &  \centering Citeseer (S) & \centering 69.80$\pm$0.02 & 69.70$\pm$0.02 &  &  \\ 

          &  \multirow{2}{=}{GraphSAINT~\cite{zeng2019graphsaint}} &  \multirow{2}{=}{\centering PyTorch-Geometric} &  \centering Cora (S) &  \centering 97.16$\pm$0.50 & 96.36$\pm$0.50  & \multirow{2}{=}{NL, AF, DR, LR, WD, Optimizer, Sampling Technique, Sample Size} &  \multirow{2}{*}{\centering \href{https://pytorch-geometric.readthedocs.io/en/latest/_modules/torch_geometric/loader/graph_saint.html}{LINK}} \\
          & &  &  \centering Citeseer (S) & \centering 91.90$\pm$0.40  & 89.66$\pm$0.30 &  &  \\ 

          &  \multirow{2}{=}{FastGCN~\cite{chen2018fastgcn}} &  \multirow{2}{=}{\centering PyTorch-Geometric} &  \centering Cora (S) & \centering 79.40$\pm$0.05 & 79.45$\pm$0.06 & \multirow{2}{=}{NL, AF, DR, LR, WD, Optimizer, Layer-wise Importance Sampling
          } &  \multirow{2}{*}{\centering \href{https://github.com/matenure/FastGCN}{LINK}} \\
          & &  &  \centering Citeseer (S) & \centering 67.70$\pm$0.09  & 66.88$\pm$0.12 &  &  \\ 

          &  \multirow{2}{=}{GAT~\cite{velickovic2017graph} } &  \multirow{2}{=}{\centering PyTorch-Geometric} & \centering Cora (S) &  \centering 81.98$\pm$0.00 & 82.00$\pm$0.01  & \multirow{2}{=}{NL, AF, DR, LR, WD, Optimizer, Attention Mechanism, nHeads, Atten. DR} &  \multirow{2}{*}{\centering \href{https://pytorch-geometric.readthedocs.io/en/latest/generated/torch_geometric.nn.conv.GATConv.html}{LINK}} \\
          & &  &  \centering Citeseer (S) & \centering 67.70$\pm$0.02  & 67.44$\pm$0.02 &  &  \\
          
          &  \multirow{2}{=}{GAT v2~\cite{brody2021attentive}} &  \multirow{2}{=}{\centering PyTorch-Geometric} & \centering Cora (S) &  \centering 80.90$\pm$0.01 & 81.00$\pm$0.02  & \multirow{2}{=}{NL, AF, DR, LR, WD, Optimizer, Attention Mechanism, nHeads, Atten. DR} &  \multirow{2}{*}{\centering \href{https://pytorch-geometric.readthedocs.io/en/latest/generated/torch_geometric.nn.conv.GATv2Conv.html}{LINK}} \\
          & &  &  \centering Citeseer (S) & \centering 67.01$\pm$0.03 & 66.30$\pm$0.04 &  &  \\ 

          &  \multirow{2}{=}{GATE~\cite{salehi2020graph}} &  \multirow{2}{=}{\centering Tensorflow} &  \centering Cora (S) &  \centering 83.10$\pm$0.02 & 83.02$\pm$0.01  & \multirow{2}{=}{NL, AF, LR, Optimizer, Lambda($\lambda$), Weight Sharing} &  \multirow{2}{*}{\centering \href{https://github.com/amin-salehi/GATE}{LINK}} \\
          & &  &  \centering Citeseer (S) & \centering 71.55$\pm$0.02 & 71.88$\pm$0.02 &  &  \\

          &  \multirow{2}{=}{H-GCN~\cite{hu2019hierarchical}} &  \multirow{2}{=}{\centering Tensorflow} &  \centering Cora (S) & \centering 82.44$\pm$0.50 & 81.98$\pm$0.50  & \multirow{2}{=}{NL, AF, LR, Optimizer, Channels, Coarsening Layers, Num. Channel} &  \multirow{2}{*}{\centering \href{https://github.com/CRIPAC-DIG/H-GCN}{LINK}} \\
          & &  &  \centering Citeseer (S) & \centering  71.84$\pm$0.60 & 70.96$\pm$0.60  &  & \\ 
            \hline
                \end{tabular}
                \label{table:opensource_nc}
            \end{table*}
        
\begin{table*}[t]
    \centering
    \caption{Comparable models from published literature using Link Prediction for evaluations. S/D: Static/Dynamic, NL: Number of Layers, AF: Activation Functions, DR: Dropout rate, LR: Learning Rate, WD: Weight Decay. }
    \begin{tabular}{m{0.5cm} m{2cm} m{1.5cm} m{2cm} m{1.5cm} m{1.5cm} m{5cm} m{0.5cm}}
        \hline
          & \centering \textbf{Model} & \centering \textbf{Platform} & \centering \textbf{Dataset} & \multicolumn{2}{c}{\centering \textbf{Metrics (\%)}} & \centering \textbf{Hyperparameters} & \textbf{Repo.} \\
        \hline
        
          \multirow{15}{*}{\rotatebox[origin=c]{90}{Link Prediction}} &
          &  &  & \vspace{0.1cm} \centering \textbf{AUC-ROC} & \vspace{0.1cm} \centering \textbf{AP}  &  & \\

          &  \multirow{2}{=}{GAE~\cite{kipf2016variational}} &  \multirow{2}{=}{\centering PyTorch} &  \centering Cora (S) &  \centering 89.55$\pm$0.02& 90.98$\pm$0.01  & \multirow{2}{=}{DR, LR, Optimizer, Encoder, Latent Space Dim., Regularization} &  \multirow{2}{*}{\centering \href{https://github.com/DaehanKim/vgae_pytorch}{LINK}} \\
          & &  &  \centering Citeseer (S) & \centering 90.53$\pm$0.02 & 91.80$\pm$0.02 &  &  \\ 

           &  \multirow{2}{=}{VGAE~\cite{kipf2016variational}} &  \multirow{2}{=}{\centering PyTorch} &  \centering Cora (S) &  \centering  89.27$\pm$0.02 & 91.24$\pm$0.01 & \multirow{2}{=}{DR, LR, Optimizer, Encoder, Regularization, Loss Terms (KL)} &  \multirow{2}{*}{\centering \href{https://github.com/DaehanKim/vgae_pytorch}{LINK}} \\
          & &  &  \centering Citeseer (S) & \centering 91.58$\pm$0.03 & 92.50$\pm$0.02 &  &  \\

          &  \multirow{2}{=}{NetGAN~\cite{bojchevski2018netgan}} &  \multirow{2}{=}{\centering Tensorflow, Pytorch} &  \centering Cora (S) &  \centering 81.81$\pm$0.00 & 85.42$\pm$0.00 & \multirow{2}{=}{Regularization, Architectures, Down-Proj, Temp Annealing, RW Length, RW Params} &  \multirow{2}{=}{\href{https://github.com/danielzuegner/netgan}{LINK},\\\href{https://github.com/mmiller96/netgan_pytorch}{LINK}} \\
          & &  &  \centering Citeseer (S) & \centering 92.33$\pm$0.00 & 92.76$\pm$0.00 &  &  \\ 

          &  \multirow{2}{=}{Linear AE~\cite{salha2021simple}} &  \multirow{2}{=}{\centering Tensorflow, PyTorch} &  \centering Cora (S) &  \centering  88.96$\pm$1.10 & 89.88$\pm$1.00 & \multirow{2}{=}{DR, LR, Epochs, Optimizer, Dimensionality d, Hidden Layer Dim.} &  \multirow{2}{=}{\href{https://github.com/deezer/linear_graph_autoencoders}{LINK},\\\href{https://github.com/jungwoonshin/Neural_Decoder_Initial}{LINK}} \\
          & &  &  \centering Citeseer (S) & \centering  90.88$\pm$1.55 & 92.44$\pm$1.78  &  &   \\ 

          &  \multirow{2}{=}{Linear VAE~\cite{salha2021simple}} &  \multirow{2}{=}{\centering Tensorflow, PyTorch} &  \centering Cora (S) &  \centering 89.77$\pm$0.98 & 90.84$\pm$0.95 & \multirow{2}{=}{DR, LR, Epochs, Optimizer, Dimensionality d, Hidden Layer Dim.} &  \multirow{2}{=}{\href{https://github.com/deezer/linear_graph_autoencoders}{LINK},\\\href{https://github.com/jungwoonshin/Neural_Decoder_Initial}{LINK}} \\
          & &  &  \centering Citeseer (S) & \centering 91.25$\pm$0.90 & 92.66$\pm$0.78 &  &  \\

          &  \multirow{2}{=}{VGRNN~\cite{hajiramezanali2019variational}} &  \multirow{2}{=}{\centering PyTorch} &  \centering Enron (D) &  \centering  92.58$\pm$0.25 & 92.66$\pm$0.50  & \multirow{2}{=}{LR, Epochs, Hidden Layer, GCN Layers, Noise Dimension, Early Stopping} & \multirow{2}{*}{\centering \href{https://github.com/VGraphRNN/VGRNN}{LINK}} \\
          & &  &  \centering Facebook (D) & \centering 87.67$\pm$0.60 & 87.88$\pm$0.70  &  &  \\ 

          &  \multirow{2}{=}{SI-VGRNN~\cite{hajiramezanali2019variational}} &  \multirow{2}{=}{\centering PyTorch} &  \centering Enron (D) &  \centering  93.14$\pm$0.50 & 93.20$\pm$0.50   & \multirow{2}{=}{LR, Epochs, Hidden Layer, GCN Layers, Noise Dimension, Early Stopping} &  \multirow{2}{*}{\centering \href{https://github.com/VGraphRNN/VGRNN}{LINK}} \\
          & &  &  \centering Facebook (D) & \centering 88.04$\pm$1.00 & 88.98$\pm$1.23 &  & \\ 
        \hline
    \end{tabular}
    \label{table:opensource_lp}
\end{table*}

\textbf{GRL-based} approaches merge reinforcement learning with graph models to selectively summarize graphs by learning from reward feedback. This method is promising for decision-centric summarization tasks like graph compression or key substructure identification. 
Designing customized deep GRL architectures for the purpose of graph summarization stands to be a promising direction in the future. However, being relatively new, GRL faces challenges in defining rewards and efficient exploration of graph spaces.

Across all approaches, we observe a trade-off between the ability to capture different aspects of graph structure and the computational efficiency. Furthermore, each method's performance can vary significantly depending on the characteristics of the dataset in use and the particular summarization task being addressed. Our analysis has revealed that the complexity of graph data and the early stage of deep learning techniques for graph mining present significant challenges. To address these hurdles and drive future progress, we have identified four potential directions for graph summarization with GNNs.

\subsection{Dynamic Graphs} 
In the real world, the data that graphs represent can evolve over time, creating changes in a graph’s topology, such as new edges that appear, nodes that disappear, or attributes that change over time. These dynamics can cause fundamental changes to the entire graph. Summarizing dynamic graphs typically involves boiling the graph down into a series of snapshots taken at various time increments. The model is then trained over these snapshots, yet a number of challenges can arise. 
To date, the approaches developed to tackle these problems have primarily focused on capturing temporal patterns and changes in graph topology. However, these methods often struggle with efficiently processing large-scale dynamic graphs and accurately capturing the evolving nature of graph relationships. Future work in this area could focus on developing more scalable algorithms that can handle larger dynamic graphs without compromising processing speed or accuracy. 

            
\subsection{Task-based Summarization} 

Graph summarization is crucial as graph sizes grow, aiding in understanding, sensemaking, and analysis. Different tasks, like detecting communities or identifying influential nodes, require tailored summarization strategies. This necessitates developing specific approaches for each unique task to effectively identify relevant patterns. Moreover, although task-based summarization is critical, this field of study has seen few successes~\cite{amiri2018netgist,amiri2020netreact} and is still an underexplored area. Open research problems include how to perform task-based summarization on streaming and heterogeneous graphs and how to leverage human feedback in the learning process.
        
\subsection{Evaluation Benchmarks} 
The optimal outcome of a graph summarization process is a ``good" summary of the original graph. However, evaluating the ``goodness" of a summary is an application-specific task that depends on the task at hand. For example, sampling-based methods are evaluated based on the quality of the sampling, while aggregation-based methods are evaluated based on the quality of classification, and so on. Current studies commonly use comparisons between their method and one or more established methods to measure the quality of their results. For instance, metrics that have been used in the literature include information loss, ease of visualization, and sparsity~\cite{liu2018graph}. However, more and different evaluation metrics are required for cases where the validation process becomes complex and more elements are involved, such as visualization and multi-resolution summaries.

\subsection{Generative Models}
In the context of generative models, graph summarization can be used to generate new graphs that have similar properties to the original ones. Generative models, such as VAEs, Graph Transformers, Graph Adversarial Networks (GANs), and Graph Auto-Regressive Models, offer effective approaches for graph summarization. These models can learn patterns in graph data and generate new graph summaries by sampling from learned distributions or sequentially generating nodes and edges. Researchers are continually exploring new techniques to pushing the boundaries of model architectures, scalability, controllability, and interpretability~\cite{bojchevski2018netgan,wickman2021sparrl,wu2020graph}. The field presents exciting opportunities for innovation and has the potential to transform various domains through more efficient and accurate graph summarization techniques.            
            
\section{Conclusion}
\label{conclusion}

New advancements in deep learning with multi-layer deep neural networks have made it possible to quickly and effectively produce a condensed representation of a large and complex graph. In this paper, we surveyed the technical trends and the most current research in graph summarization with GNNs. We provided an overview of different graph summarization techniques and categorized and described the current GNN-based approaches of graph summarization. We also discussed a new line of research focusing on RL methods to evaluate and improve the quality of graph summaries. 

To advance research in this field, we also outlined several frequently-used benchmarking tools, including datasets, open-source codes, and techniques for generating summarized graphs. In addition, we identified four promising directions for future research based on our findings from the survey. We strongly believe that using GNNs for graph summarization is not just a passing trend. Rather, it has a bright future in a wide range of applications across different domains. 

As a potential area of focus for our future work, we endeavor to delve into the capabilities of GNN-based generative models including VGAEs~\cite{kipf2016variational} and GANs~\cite{bojchevski2018netgan,wu2020graph}, to push the boundaries of graph summarization and generation. Additionally, we will explore the potential of GRL~\cite{wickman2021sparrl} to create new graphs based on their summarized representations. By addressing challenges and expanding the frontiers of graph synthesis, we envision empowering data analysts and researchers with powerful tools for efficient and insightful analysis of complex graph-structured data.


\bibliographystyle{IEEEtran}
\bibliography{references}

\begin{IEEEbiography}[{\includegraphics[width=1in,height=1.25in,clip,keepaspectratio]{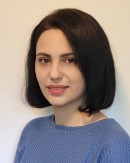}}]{Nasrin Shabani}{\space}received a Master of Research degree in Computer Science with First Class Honours from the Macquarie University, Sydney, NSW, Australia. She is currently pursuing a Ph.D. in Computer Science at the same institution. Her research interests lie at the intersection of graph mining, graph summarization, and deep learning. Through her work, she aims to develop novel algorithms and techniques that can extract meaningful insights and patterns from complex graph data structures.
\end{IEEEbiography}

\begin{IEEEbiography}[{\includegraphics[width=1in,height=1.25in,clip,keepaspectratio]{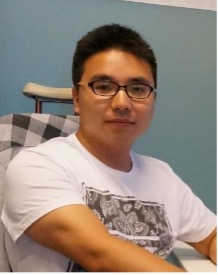}}]{Jia Wu} (Senior Member, IEEE) is currently the Research Director of the Centre for Applied Artificial Intelligence and the Director of Higher Degree Research in the School of Computing at Macquarie University, Sydney, Australia. Dr Wu received his Ph.D. degree in computer science from the University of Technology Sydney, Australia. His current research interests include data mining and machine learning. Since 2009, he has published 100+ refereed journal and conference papers, including TKDE, TKDD, KDD, ICDM, WWW, and NeurIPS.
\end{IEEEbiography}

\begin{IEEEbiography}[{\includegraphics[width=1in,height=1.25in,clip,keepaspectratio]{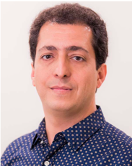}}]{Amin Beheshti}{\space}holds B.S. (1st Hons.) and M.S. degrees (1st Hons.) in computer science and engineering, and a Ph.D. in computer science from UNSW Sydney, Australia. Amin is a Full Professor of data science at Macquarie University. He is currently the Director of the Centre for Applied Artificial Intelligence and the Head of the Data Science Research Laboratory, School of Computing, Macquarie University. He is a leading author of several authored books in data, social, and process analytics, co-authored with other high-profile researchers.

\end{IEEEbiography}

\begin{IEEEbiography}[{\includegraphics[width=1in,height=1.25in,clip,keepaspectratio]{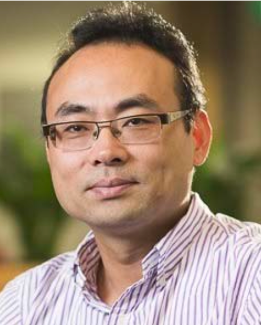}}]{Quan Z. Sheng}{\space}received his Ph.D. degree in computer science from the University of New South Wales, Sydney, NSW, Australia.
He is currently a full Professor and Head of the School of Computing, at Macquarie University, Sydney. His research interests include big data analytics, service-oriented computing, and the Internet of Things. 
Microsoft Academic ranked Prof. Michael Sheng as one of the Most Impactful Authors in Services Computing (ranked Top 5 All-Time) and in the Web of Things (ranked Top 20 All-Time). 
\end{IEEEbiography}

 \begin{IEEEbiography}[{\includegraphics[width=1in,height=1.25in,clip,keepaspectratio]{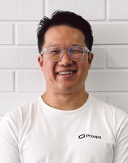}}]{Jin Foo}{\space}is the Staff Data Scientist at Prospa, Australia's top online lender to small businesses and was previously Data Science Lead at Woolworths Group; specialising in identity resolution, hyper-personalised offers, propensity modelling, sequential bin packing and time-series analysis. Jin is currently a 2nd year Master of Research student at the School of Computing, Macquarie University, Sydney, Australia. His research focuses on anomaly detection with word embeddings, hashing algorithms and graph networks. 
\end{IEEEbiography}

 \begin{IEEEbiography}[{\includegraphics[width=1in,height=1.25in,clip,keepaspectratio]{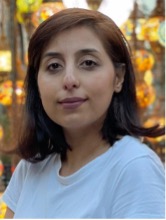}}]{Venus Haghighi}{\space}is currently a Ph.D. student in computer science at the School of Computing, Macquarie University, Sydney, NSW, Australia. The focus of her research is to enhance classic GNN models and explore robust graph learning paradigms to detect and mitigate the camouflage behavior of malicious actors in both static and dynamic networks. Her research interests include graph-based anomaly detection, graph neural networks, graph-based fraud detection, and graph data mining. 
\end{IEEEbiography}

\begin{IEEEbiography}
[{\includegraphics[width=1in,height=1.25in,clip,keepaspectratio]{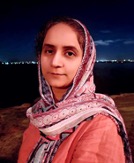}}]{Ambreen Hanif}{\space}is a 2nd year Ph.D. student at the School of Computing, Macquarie University NSW, Australia. After completing her Master's degree, she decided to continue her research as a Ph.D. candidate. Her research interests lie in the field of Explainable Artificial Intelligence (XAI) for Deep Neural Networks, Data Provenance, and Storytelling with XAI. Specifically, she aims to develop novel methods to enhance the interpretability and transparency of deep neural networks. 
\end{IEEEbiography}

\begin{IEEEbiography}[{\includegraphics[width=1in,height=1.25in,clip,keepaspectratio]{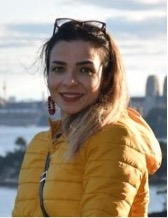}}]{Maryam Shahabikargar}{\space}received her Master of Research degree in computer science from the Macquarie University, Sydney, Australia. After completing her Master's degree, she decided to continue her research as a Ph.D. candidate in computer science at Macquarie University. Her current research interests include not only NLP and graph embeddings but also link prediction and anomaly detection. She aims to develop a model that combines her research interests to tackle a specific problem in the field of finance. 

\end{IEEEbiography}
\end{document}